\pdfoutput=1

\documentclass[11pt]{article}
\usepackage{amsmath}
\usepackage{booktabs}

\usepackage[final]{acl}
\usepackage{longtable}
\usepackage{times}
\usepackage{latexsym}
\usepackage{subcaption} 
\usepackage{float}
\usepackage{multirow} 
\usepackage{comment}
\usepackage{enumitem} 
\usepackage[T1]{fontenc}

\usepackage[utf8]{inputenc}

\usepackage{microtype}

\usepackage{inconsolata}

\usepackage{graphicx}

\title{Scaling Performance and Low-Resource Annotation with Many-Shot In-Context Learning for Named Entity Recognition}

\author{\textbf{Qi Zhang}$^1$ \quad
        \textbf{Fangping Lan}$^1$ \quad \\
        \textbf{Cornelia Caragea}$^2$ \quad
        \textbf{Longin Jan Latecki}$^1$ \quad
        \textbf{Eduard Dragut}$^1$\\
  $^1$Temple University\quad $^2$University of Illinois Chicago\\ 
  {\tt \{qi.zhang, latecki, edragut\}@temple.edu,  cornelia@uic.edu} \\
  }

\begin{document}
\maketitle
\begin{abstract}
In-context learning (ICL) with large language models (LLMs) has 
emerged as a powerful alternative to fine-tuning for Named Entity Recognition (NER), achieving strong performance with minimal annotation and no additional training.
However, prior work has shown that despite their adaptability, LLMs still lag behind fully supervised models such as fine-tuned BERT in structured tasks like NER.
While existing studies on ICL for NER have mainly explored few-shot settings, the potential of scaling to hundreds of demonstrations has not been thoroughly investigated.
To address this gap, we conduct a comprehensive investigation of many-shot ICL for NER and further explore its effectiveness in annotating and refining data for low-resource NER tasks. Specifically, we evaluate various LLMs across multiple domains using hundreds of ICL examples and then assess the feasibility of using many-shot ICL as a data annotation framework.
Our experiments demonstrate that: (1) scaling to hundreds of in-context examples enables LLMs to match or even surpass the performance of fully supervised BERT models; and (2) using about one hundred human-labeled examples as demonstrations, many-shot in-context annotation can generate high-quality labeled data, leading to approximately 10\% absolute F1 improvement over existing state-of-the-art approaches when used to fine-tune BERT on low-resource NER \footnote{Code and data are available at \href{https://github.com/TUDMLab/Many-Shot-ICL-NER}{Many-Shot-ICL-NER}.}.
\end{abstract}

\section{Introduction}
Named Entity Recognition (NER) is a fundamental task in information extraction that aims to identify and classify entities such as persons, organizations, locations, and domain-specific concepts within text \cite{jurafsky2000speech,pan-etal-2025-taxonomy}. 
It serves 
various downstream applications, including knowledge base and knowledge graph construction \cite{zhang-etal-2024-scier, wei-etal-2024-collabkg}, question answering \cite{choudhary-etal-2023-iitd}, and domain-specific document understanding \cite{singh-etal-2023-scirepeval, nandi-agrawal-2024-improving}. 
State-of-the-art (SOTA) NER are small models such as BERT fine-tuned on in-domain data \cite{hu2024three,zhang2024cross,zhang2022exploring}. They require large amount of human-labeled examples \cite{devlin-etal-2019-bert,ding-etal-2021-nerd,sun2025positionalattentionefficientbertbased}. A typical workaround is to generate large-scale labeled instances using rule-based or heuristic functions \cite{pryzant-etal-2022-automatic,Safranchik_Luo_Bach_2020,shang-etal-2018-learning}, followed by fine-tuning with a smaller set of  human-labeled samples. 
In this work, we explore the use of large language models (LLMs) for in-context learning (ICL) to perform NER without updating model parameters \cite{wang-etal-2025-gpt, ashok2023promptner,pan-etal-2025-climateie}. Our key insight is that with a moderate number of human-annotated examples (e.g., 100), LLMs can act as in-context annotation functions, generating thousands of high-quality, labeled samples for training smaller, more efficient models.

While LLMs in few-shot settings still underperform compared to fully supervised small models \cite{xie-etal-2024-self, heng-etal-2024-proggen} and are slower at inference \cite{zaratiana-etal-2024-gliner, bogdanov-etal-2024-nuner}, they excel at generalizing from a few examples, making them ideal for bootstrapping annotations in low-resource domains. Rather than relying on LLMs as full-time NER systems, we leverage them to generate training data for more efficient small models. 
This approach is made feasible by the increased context length in modern LLMs, which supports many-shot ICL—scaling the number of demonstrations from a few to several hundred \cite{fu2024data, gao2024train}. Prior work shows that ICL performance improves with more demonstrations \cite{bertsch-etal-2025-context, song-etal-2025-many, agarwal2024many}, yet many-shot ICL paradigm remains underexplored in NER.

We propose an In-Context Annotation (ICA) framework that leverages many-shot ICL to automatically label domain-specific NER datasets. In low-resource settings, ICA uses 100 human-labeled examples to annotate 2k unlabeled instances. To enhance annotation quality, we apply refinement strategies such as self-consistency, self-correction, and error-aware self-refinement (ERA). 
The annotated data is then used to fine-tune a BERT-based model, which outperforms baselines by up to 10 F1 points—achieving competitive results with substantially less human effort.

We conduct a comprehensive study of many-shot ICL for NER across four domains, evaluating both open-source and closed-source LLMs. Our experiments show that using around 100 in-context examples allows LLMs to match or even surpass the performance of supervised models trained on thousands of human-annotated instances, demonstrating the data efficiency of the ICA framework. 
Finally, we analyze the scalability of ICA, the impact of seed set size, and the potential of using smaller or open-source models for annotation. These findings offer practical insights into the strengths and limitations of LLMs as efficient NER annotators, guiding future research and deployment.
To sum up, our contributions are four-fold:
\begin{itemize}[noitemsep, topsep=0pt, partopsep=0pt, parsep=0pt, leftmargin=*]
  \item \textbf{Empirical Study.} We present the first systematic study of \emph{many-shot} in-context learning (ICL) for NER, showing consistent, monotonic gains as the number of demonstrations increases across four datasets and seven LLMs (\S\ref{sec:manyshot-icl}).
  \item \textbf{Framework.} We propose a \emph{many-shot In-Context Annotation (ICA)} framework that relies on a modest number of human-labeled samples. ICA includes several refinement strategies to enhance annotation quality (\S\ref{sec:ica_framework}).
  \item \textbf{Experiments.} Through experiments on five low-resource NER datasets, we show that our approach outperforms SOTA baselines by a substantial margin, achieving an average absolute F1 improvement of approximately 10 points (\S\ref{sec:ica_results}).
  \item \textbf{Analyses.} We conduct extensive analyses on the scalability
of ICA, the efficiency of seed annotation, and the feasibility of using smaller, open-source LLMs for annotation tasks (\S\ref{sec:discussion}).
\end{itemize}

\section{Many-Shot In-Context Learning NER} \label{sec:manyshot-icl}
In this section, we first describe the task formulation of ICL for NER in \S \ref{sec:ner_task},
and then present our experimental setup and results in \S\ref{sec:micl_ner}.

\subsection{In-Context Learning for NER} \label{sec:ner_task}

We formulate the NER task as a conditional generation problem. Given an input sentence, the goal is to generate its annotated version in a structured format where entity spans and type are explicitly marked. 
This formulation enables the use of LLMs via ICL, allowing the model to learn the labeling behavior from a set of demonstration examples and apply it to test instances.
Let \( \mathcal{D}_{\text{train}} \) and \( \mathcal{D}_{\text{test}} \) denote the training and test datasets, respectively. We construct the ICL demonstration set by sampling \( k \) labeled examples from the training set:
\begin{equation}
\mathcal{D}_{\text{ICL}} = \{(x_1, y_1), (x_2, y_2), \dots, (x_k, y_k)\} \subset \mathcal{D}_{\text{train}}
\label{eq:icl_demo_set}
\end{equation}
where \( x_i \) is an input sentence and \( y_i \) is its output in the target format.
Given a test input \( x_{\text{test}} \in \mathcal{D}_{\text{test}} \), we construct the prompt as:
\begin{equation}
\texttt{Prompt} = \texttt{I} + \sum_{i=1}^{k} \texttt{Format}(x_i, y_i) + \texttt{Format}(x_{\text{test}})
\label{eq:icl_prompt}
\end{equation}
where \( \texttt{I} \) denotes the task instruction, and \( \texttt{Format}(x, y) \) renders a labeled example in a consistent input-output style.
The model then generates:
\begin{equation}
\hat{y} = \text{LLM}(\texttt{Prompt})
\label{eq:icl_generation}
\end{equation}
which aims to recover the entity-annotated version of \( x_{\text{test}} \). Formally, we model the generation of the output as a conditional probability over the instruction, the demonstration set, and the test input:
\begin{equation}
\hat{y}_{\text{test}} = \arg\max_{y} \; P(y \mid \texttt{I}, \mathcal{D}_{\text{ICL}}, x_{\text{test}})
\label{eq:icl_objective}
\end{equation}

\begin{figure}[t]
\centering
\includegraphics[width=\linewidth]{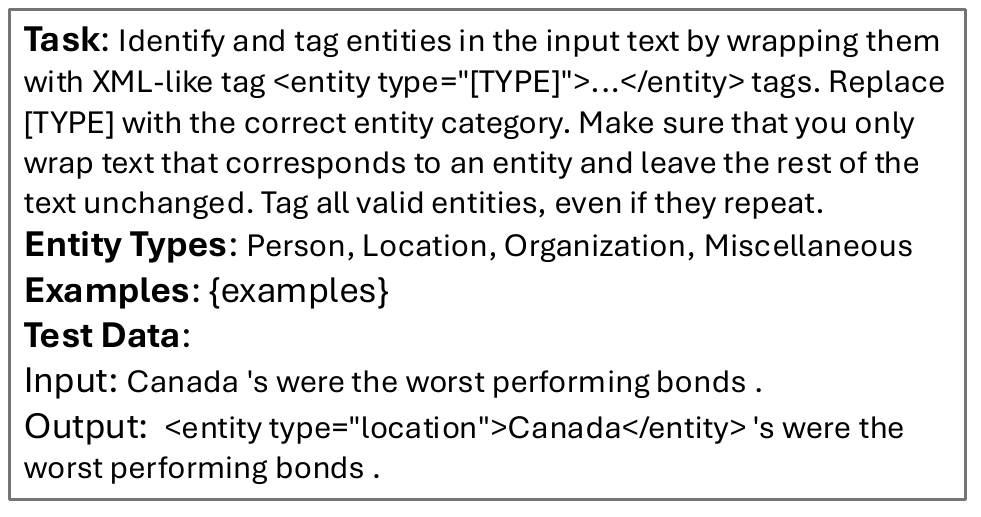}
\vspace{-17pt}
\caption{
XML-style tagging prompt. \texttt{\{examples\}} denotes the in-context demonstrations, each following the same input-output format.
}
\label{fig:prompt}
\vspace{-10pt}
\end{figure}

\paragraph{XML-style Output Format.}
To represent both entity spans and their types in generation, we adopt an XML-style annotation format in which each entity mention is wrapped with explicit tags that include the entity type as an attribute. 
Figure~\ref{fig:prompt} shows the prompt used in our ICL for NER.
This XML-style generation format, which explicitly marks both entity boundaries and types, and preserves surrounding non-entity tokens, has been shown to improve the boundary sensitivity of LLMs and reduce span-level annotation errors \cite{ding-etal-2024-rethinking, zhang-etal-2024-scier, hu2024improving}. 
Moreover, the structured output can be easily converted into the traditional Beginning-Inside-Outside (BIO) format, allowing direct comparison with results from previous NER systems. 
We adopt this XML-style format as the default prompting template throughout this work.

\begin{figure*}[t]
  \centering
  \begin{minipage}[t]{0.245\textwidth}
    \centering
    \includegraphics[width=\linewidth]{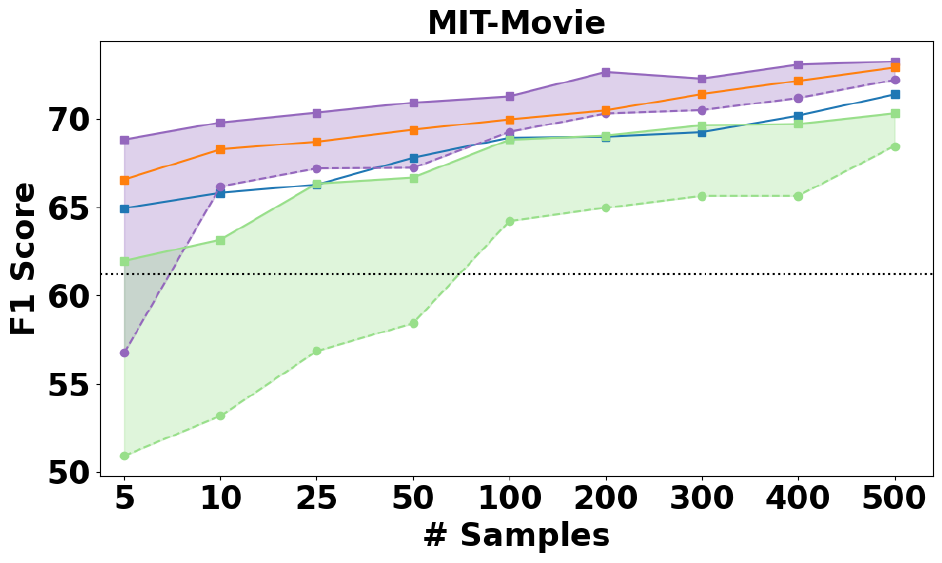}
    \vspace{-8pt}
  \end{minipage}
  \hfill
  \begin{minipage}[t]{0.245\textwidth}
    \centering
    \includegraphics[width=\linewidth]{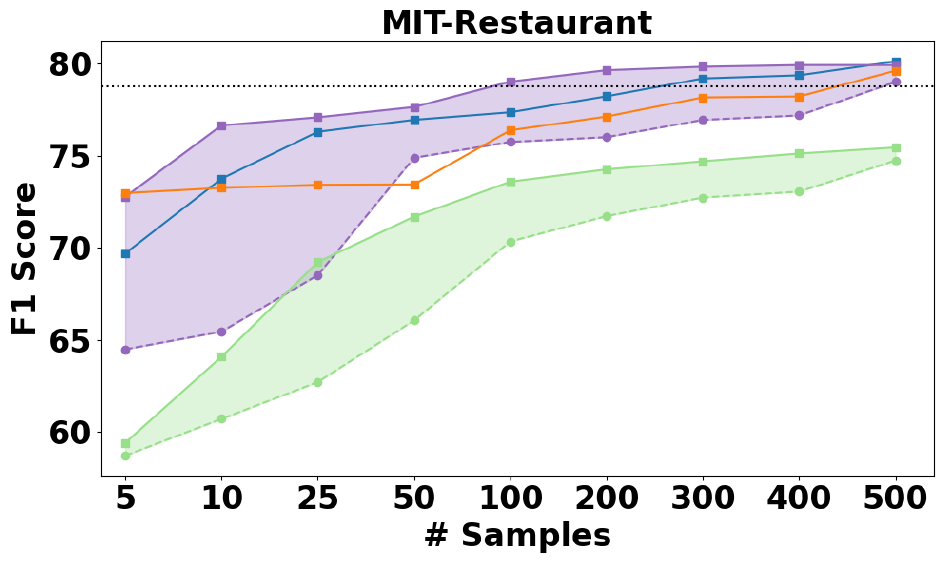}
    \vspace{-8pt}
  \end{minipage}
  \hfill
  \begin{minipage}[t]{0.245\textwidth}
    \centering
    \includegraphics[width=\linewidth]{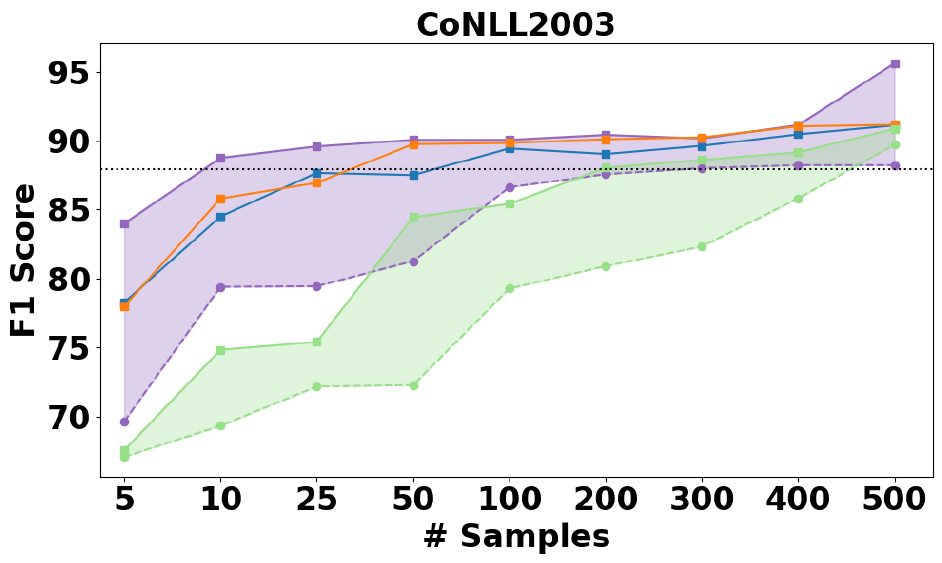}
    \vspace{-8pt}
  \end{minipage}
  \hfill
  \begin{minipage}[t]{0.245\textwidth}
    \centering
    \includegraphics[width=\linewidth]{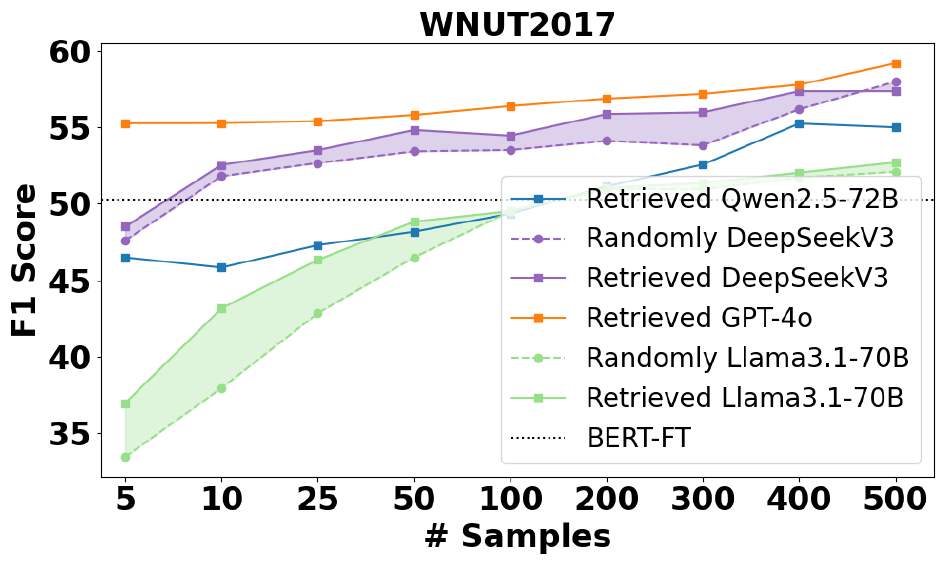}
    \vspace{-8pt}
  \end{minipage}
  \vspace{-15pt}
  \caption{Performance trends on the MIT-Movie, MIT-Restaurant, CoNLL2003 and WNUT2017 datasets as the number of samples increases.}
  \label{fig:many_shots_trends}
\vspace{-10pt}
\end{figure*}

\subsection{Experiments} \label{sec:micl_ner}

\paragraph{Datasets.}
To ensure diversity and alignment with real-world application scenarios, we evaluate many-shot ICL on four NER datasets from different domains: MIT-Restaurant, MIT-Movie \cite{mitner}, WNUT 2017 \cite{derczynski-etal-2017-results}, and CoNLL2003 \cite{tjong-kim-sang-de-meulder-2003-introduction}.
CoNLL2003 is a widely used benchmark for standard NER tasks in general domains. 
MIT-Restaurant and MIT-Movie are domain-specific NER datasets with 8 and 12 fine-grained entity types, respectively, posing a greater challenge due to their specialized vocabulary and entity distribution.
WNUT 2017 focuses on NER in noisy and informal social media texts, further testing a model’s robustness under real-world conditions.
More dataset details are provided in Appendix~\ref{apx:micl_dataset}.

\paragraph{Models and Settings.}
We evaluate a diverse set of LLMs, including open-source models such as Qwen-2.5 (7B, 32B, and 72B) \cite{qwen} and LLaMA-3.1 (8B and 72B) \cite{llama}, as well as frontier models like GPT-4o \cite{gpt4o} and DeepSeekV3 \cite{deepseekv3}. 
All models support a context length of at least 32K tokens, which enables them to accommodate hundreds of in-context examples.
For each dataset, we gradually increase the number of demonstrations up to 500 to study the scaling behavior of many-shot ICL.
To select in-context examples, we consider two strategies: (1) \textbf{random sampling}: randomly sampling $k$ examples from the training set as fixed demonstrations for the entire test set; and (2) \textbf{retrieval-based}: selecting the top-$k$ training instances most similar to each test sample based on lexical similarity using BM25.
For the supervised baseline, we fine-tune a BERT-based model on the full training set of each dataset, referred to as \textbf{BERT-FT}. 
We additionally include a target-only variant, \textbf{BERT-FT (target-only)}, which is fine-tuned solely on the limited human-annotated examples available in each low-resource target domain (100--200 samples), without any access to source-domain data. This variant serves as a lower-bound reference that isolates the contribution of our ICA-based data generation from any source-domain transfer.
To assess the impact of LLM choice on downstream annotation quality, we compare annotations generated by GPT-4o, DeepSeekV3, and three variants of Qwen-2.5 (7B, 32B, 72B) in our analysis (\S\ref{sec:discussion}).
All reported scores are entity-level micro-F1 with exact span matching, computed using the \texttt{seqeval} package under default settings.
Appendix~\ref{apx:micl_exp_details} describes the experimental settings.


\paragraph{Results.}

Due to page constraints, we present a subset of our results in Figure~\ref{fig:many_shots_trends}. 
Comprehensive experimental results and detailed score tables are in Appendix~\ref{apx:micl_ner_detail}.
Across all domains, we observe a consistent improvement in F1-score as the number of in-context demonstrations increases. 
This trend is especially evident for the retrieval-based ICL setting, where performance improves rapidly and plateaus around 100 examples. 
An exception is observed on CoNLL2003, where performance saturates even earlier (around 50 examples). 
In contrast, randomly sampled demonstrations yield a more gradual and steady improvement as the number of examples grows. 
Notably, for the same model, the performance gap between retrieval-based and random sampling narrows as more examples are included, with the two approaches showing comparable results when scaled to 500-shot.

While retrieval-based ICL consistently outperforms random sampling in most cases (except WNUT2017 with 500 samples), it is important to highlight the cost difference between the two settings. 
Random sampling selects a fixed set of $k$ demonstrations once and reuses them for all test examples. 
In contrast, retrieval-based ICL involves dynamically selecting top-$k$ nearest neighbors for each test instance from the training pool, resulting in significantly higher annotated data reqiurement. 
Our results suggest that although random sampling lags behind in performance initially, increasing the number of random samples can yield highly competitive results, approaching those of the more expensive retrieval-based approach.

Comparing the many-shot ICL performance against BERT-FT, we found that in nearly all cases, except for LLaMA3.1-70B on MIT-Restaurant, LLMs achieve performance comparable to or even exceeding that of BERT-FT on the full training set.
This highlights the strong data efficiency of many-shot ICL, demonstrating that LLMs can match supervised performance using only hundreds of labeled demonstrations without additional parameter updates.
This finding is especially important for random sampling, which achieves competitive performance while requiring much less annotation effort than both BERT-FT and retrieval-based ICL.
For more results and complete per-model evaluation tables, refer to Appendix~\ref{apx:micl_ner_detail}.

\section{Many‑Shot In‑Context Annotation for Low-Resource NER} \label{sec:ica}




The results in Section~\ref{sec:manyshot-icl} show that many-shot in-context learning enables LLMs to achieve strong performance in NER with only around hundreds of labeled examples.
However, despite its effectiveness, many-shot ICL requires significant inference cost—each new prediction involves long prompts and repeated calls to large models.
This motivates a practical alternative: instead of relying on LLMs for expensive real-time inference, we consider employing them as annotators to generate high-quality labeled data that can be used to train smaller, more efficient models like BERT.
Motivated by this, we hypothesize that if a small number of labeled examples can guide high-quality ICL predictions, then the same setup can be used to automatically annotate unlabeled corpora.
We consider a realistic setting in which only a modest set of human-annotated examples (e.g., 100) is available.
We aim to use these examples as fixed in-context demonstrations to prompt an LLM and annotate thousands of unlabeled sentences.
We refer to this process as \textbf{Many-shot In-Context Annotation (ICA)}, where a fixed set of human-labeled examples is used as in-context demonstrations to prompt an LLM for annotating large-scale unlabeled data.

In this section, we first introduce our proposed ICA framework in \S\ref{sec:ica_framework}, which generates structured annotations for each sentence using the same fixed prompt, enabling the training of a downstream BERT-based NER model.
To further enhance annotation quality, we propose three strategies: \textbf{self-consistency}, \textbf{self-correction}, and a \textbf{error-aware refinement} (ERA) method aimed at correcting common NER-specific annotation errors.
We describe the experimental setup in \S\ref{sec:ica_exp_setting}, followed by results across five low-resource domains in \S\ref{sec:ica_results}.

\subsection{ICA Framework} \label{sec:ica_framework}

\begin{figure}[!]
\centering
\includegraphics[width=\linewidth]{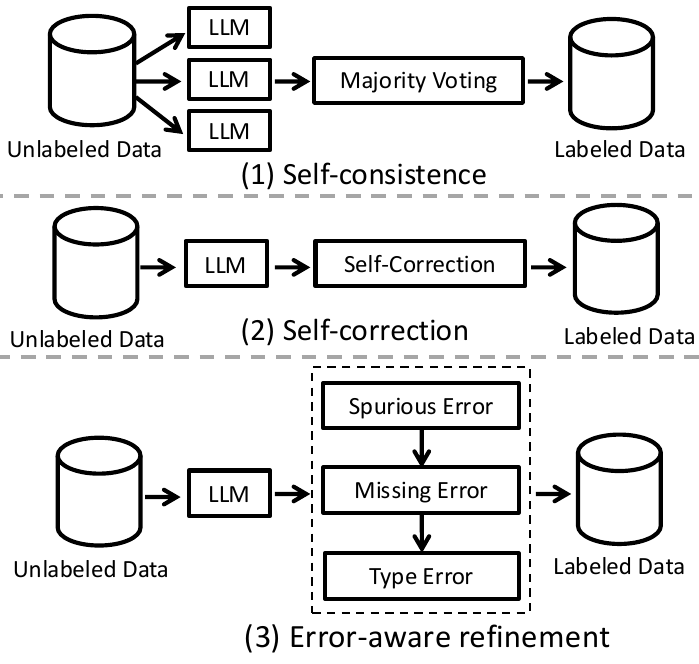}
\vspace{-20pt}
\caption{
Three refinement methods used in our proposed ICA framework.
}
\label{fig:ica_pipeline}
\vspace{-10pt}
\end{figure}

We describe here our ICA framework,
which builds upon the ICL formulation introduced in \S~\ref{sec:ner_task}. 
Given a small set of human-annotated examples, we use them as fix demonstrations to prompt an LLM and annotate a much larger unlabeled corpus.
The generated annotations follow the same XML-style output format as described in \S~~\ref{sec:ner_task}, and are converted to BIO format to train a downstream BERT-based NER model.
To further improve the quality of annotated data, we introduce a refinement step based on LLM confidence.
Specifically, we calculate the average log probability over the entity tokens enclosed by XML tags, excluding all non-entity tokens. 
Formally, for a generated sentence $\hat{y}$ annotated by the LLM by Eq. \ref{eq:icl_generation} to include entity spans $\{s_1, s_2, \dots, s_n\}$, where each $s_i$ consists of a token sequence representing an entire XML-style entity annotation (e.g., \texttt{<entity type="type">mention</entity>}), we define the confidence score as:
\begin{equation}
\label{eq:confidence}
\text{Confidence}(\hat{y}) = \frac{1}{\sum_{i=1}^{n} |s_i|} \sum_{i=1}^{n} \sum_{t \in s_i} \log p(t)
\end{equation}
where $|s_i|$ denotes the number of tokens in the $i$-th entity span, and $p(t)$ is the generation probability of token $t$.
Consider the example sentence in Figure~\ref{fig:prompt}, the confidence score is computed using the log probabilities of the entire entity span \texttt{<entity type="location">Canada</entity>}.
This provides a more focused estimate of annotation confidence.



We begin with a standard ICA setup, where each unlabeled sentence is annotated via a single LLM call using a fixed prompt with human-labeled demonstrations.
To improve annotation quality, we further introduce three refinement strategies: self-consistency, self-correction, and error-aware refinement (EAR). 
They are applied to the bottom 50\% of low-confidence samples obtained by Eq. \ref{eq:confidence}.
As shown in Figure~\ref{fig:ica_pipeline}, each strategy builds on the initial ICA output in different ways:
(1) Self-consistency performs two additional LLM predictions for each low-confidence sentence, using varied sampling and demonstration orders, and aggregates the three outputs via majority voting over both entity spans and types.
(2) Self-correction introduces a second prompting step where the LLM is explicitly asked to review and revise its own annotations.
(3) EAR applies three dedicated correction prompts in sequence to handle common NER-specific errors, including spurious entities, missing entities, and incorrect entity types.
Due to space constraints, detailed implementation and prompt designs are  in Appendix~\ref{apx:ica_refinement}.

\subsection{Experimental Settings} \label{sec:ica_exp_setting}
\paragraph{Dataset.}
We conduct experiments on the CrossNER \cite{crossner} benchmark, which covers five diverse domains: \textit{AI}, \textit{Literature}, \textit{Music}, \textit{Politics}, and \textit{Science}. 
We choose CrossNER as our testbed because it presents a realistic and challenging low-resource scenario: each domain contains a large number of entity types but only 100 or 200 labeled examples. 
Moreover, the small size of the labeled data reduces the risk of data contamination from LLM pretraining corpora \cite{deng-etal-2024-investigating}, making it a cleaner and more trustworthy benchmark for evaluating low-resource NER methods.
For each domain, we randomly sample 100 human-annotated sentences from the provided training set to serve as our in-context demonstration set. These examples are used to prompt the LLM for annotation. 
Detailed statistics of each domain are provided in Appendix~\ref{apx:corssner}.


\paragraph{Unlabeled Data Collection.}
Inspired by prior retrieval-augmented data augmentation methods \cite{cai-etal-2023-graph}, we use BM25 to retrieve the top-10 most similar unlabeled sentences for each of the 100 seed examples.
This yields a set of 1k sentences that are semantically close to the labeled data and likely to contain relevant entity contexts.
To further improve coverage and diversity, we additionally sample 1k sentences uniformly at random from the remaining unlabeled corpus (excluding the retrieved sentences).
The unlabeled corpus in each domain is sourced from the CrossNER benchmark \cite{crossner}, which provides domain-specific Wikipedia sentences.
This setup is consistent with prior data augmentation work that also leverages Wikipedia-derived corpora \cite{cai-etal-2023-graph}.
The final annotation set for each domain therefore contains 2k unlabeled sentences, combining high-relevance and high-diversity examples.

\paragraph{Evaluation.}
We use the DeepSeekV3 
to perform annotation over the 2k unlabeled sentences collected as described above. 
The generated annotations are converted into the standard BIO format and used to train a BERT-based NER model.
To evaluate the effectiveness of the proposed ICA, we compare our approach with several representative baselines from two categories. 
The first group includes traditional data augmentation methods such as DAGA \cite{ding-etal-2020-daga}, NERDA \cite{dai-adel-2020-analysis}, and GPDA \cite{cai-etal-2023-graph} and LLM-based data augmentation methods ProgGen\cite{heng-etal-2024-proggen}. 
The second group includes recent state-of-the-art methods for low-resource or cross-domain NER, such as 
LST-NER \cite{zheng-etal-2022-cross}, DoSEA \cite{tang-etal-2022-dosea}, \citet{hu-etal-2022-label}, DTrans-SMix \cite{zhang2022exploring}, DH-GAT \cite{xu2023decoupled}, Three-KPNs \cite{hu2024three}, DTrans-MPrompt \cite{zhang2024cross}, PromptNER\cite{ashok2023promptner}, IF-WRANER \cite{nandi-agrawal-2024-improving}, GliNER~\cite{zaratiana-etal-2024-gliner}, and B2NER~\cite{yang-etal-2025-beyond}.
We also include the performance of using LLM to do in-context inference with zero-shot (ICL-ZS) and 100-shot (ICL-MS) for comparision.
More details about the baselines implement are provided in Appendix~\ref{apx:baselines}.

\subsection{Results} \label{sec:ica_results}

\begin{table*}[t]
\centering
\small
\setlength{\tabcolsep}{5pt}
\begin{tabular}{l|ccccc|c}
\toprule
\textbf{Method} & \textbf{AI} & \textbf{Literature} & \textbf{Music} & \textbf{Politics} & \textbf{Science} & \textbf{Average} \\
\midrule
\multicolumn{7}{l}{\textbf{LLM In-Context Inference}} \\
ICL-ZS & 59.68 & 61.49 & 49.03 & 72.25 & 68.44 & 62.18 \\
ICL-MS & 74.82 & 77.83 & 83.72 & 83.05 & 77.79 & 79.44 \\
ICL-MS w/ Self-Correction & 76.15 & 78.92 & 84.65 & 83.94 & 78.85 & 80.50 \\
ICL-MS w/ EAR & 77.95 & 80.35 & 85.68 & 85.15 & 80.12 & 81.85 \\
\midrule
\multicolumn{7}{l}{\textbf{State-of-the-art Approaches}} \\
LST-NER  & 63.28 & 70.76 & 76.83 & 73.25 & 70.07 & 70.84 \\
DoSEA  & 66.03 & 68.59 & 73.10 & 71.69 & 75.52 & 70.99 \\
DTrans-SMix  & 68.93 & 69.22 & 76.10 & 76.70 & 72.35 & 72.66 \\
DH-GAT  & 69.30 & 72.51 & 78.77 & 77.06 & 74.21 & 74.37 \\
B2NER       & 59.00 & 63.70 & 68.60 & 67.80 & 72.00 & 66.22 \\
GLiNER$^\dagger$ & 70.55 & 75.13 & 83.27 & 81.11 & 75.73 & 77.16 \\
Three-KPNs  & 70.05 & 70.39 & 77.57 & 77.82 & 73.63 & 73.89 \\
PromptNER  & 68.83 & 74.44 & 84.26 & 78.61 & 72.59 & 74.95 \\
DTrans-MPrompt  & 70.13 & 73.51 & 79.54 & 80.54 & 73.06 & 75.36 \\
IF-WRANER  & 68.81 & 75.52 & 85.43 & 75.31 & 79.80 & 76.97 \\
\midrule
\multicolumn{7}{l}{\textbf{Data Augmentation Approaches}} \\
DAGA  & 66.77 & 71.15 & 78.48 & 73.30 & 73.07 & 72.55 \\
NERDA  & 70.20 & 71.28 & 79.56 & 75.30 & 74.37 & 74.14 \\
GPDA   & 70.05 & 72.34 & 80.16 & 75.95 & 75.55 & 74.81 \\
ProgGen     & 71.56 & 73.31 & 85.14 & 84.19 & 78.25 & 78.49 \\
\midrule
\multicolumn{7}{l}{\textbf{LLM-assisted In-Context Annotation (Ours)}} \\
ICA & 76.97$_{\pm 0.42}$ & 80.95$_{\pm 0.38}$ & 87.95$_{\pm 0.29}$ & 87.50$_{\pm 0.35}$ & 81.71$_{\pm 0.41}$ & 83.01$_{\pm 0.37}$ \\
ICA w/ Self-Consistency & 77.40$_{\pm 0.36}$ & 82.03$_{\pm 0.31}$ & 88.51$_{\pm 0.25}$ & 88.22$_{\pm 0.28}$ & 82.35$_{\pm 0.33}$ & 83.70$_{\pm 0.31}$ \\
ICA w/ Self-Correction & 77.79$_{\pm 0.28}$ & 81.89$_{\pm 0.25}$ & 89.53$_{\pm 0.22}$ & 87.69$_{\pm 0.26}$ & 82.02$_{\pm 0.29}$ & 83.78$_{\pm 0.28}$ \\
\textbf{ICA w/ EAR} & \textbf{79.88}$_{\pm 0.21}$ & \textbf{82.96}$_{\pm 0.18}$ & \textbf{89.30}$_{\pm 0.15}$ & \textbf{88.69}$_{\pm 0.19}$ & \textbf{83.17}$_{\pm 0.23}$ & \textbf{84.80}$_{\pm 0.18}$ \\
\bottomrule
\end{tabular}
\vspace{-5pt}
\caption{F1 scores on CrossNER. Results for our ICA methods are averaged over five runs with different random seeds; subscripts denote standard deviation. $^\dagger$GLiNER is fine-tuned with \texttt{gliner\_large-v2.5} on 100 human-annotated examples per domain using the official training script.}
\label{tab:main-results}
\end{table*}

Table~\ref{tab:main-results} presents the performance of our proposed ICA-based annotation strategies, compared with multiple strong baselines, including LLM-based inference, traditional data augmentation methods, and recent SOTA low-resource NER approaches. We report F1-scores averaged over five runs across five domains from the CrossNER benchmark.


In summary, our results provide strong evidence that \textbf{LLM-assisted In-Context Annotation} substantially outperforms both traditional data augmentation techniques and SOTA low-resource NER methods. 
Importantly, our framework relies solely on 100 human-annotated examples per domain, whereas prior methods typically require additional development sets and access to source-domain annotated corpora. 
\textit{These findings highlight the power of pre-trained LLMs and their in-context learning capabilities for efficient and scalable dataset construction in low-resource settings.} 
They suggest that combining a small amount of acceptable human supervision with powerful LLMs via in-context learning can serve as a practical and scalable paradigm for building NER datasets in resource-constrained domains.


Compared to direct in-context inference using LLMs (ICL-MS), our ICA-based approach offers both higher performance and greater practical utility. While ICL-MS achieves strong results (79.44 average F1), our best ICA variant (ICA with EAR) improves this by over 5 F1-scores, indicating that converting LLM predictions into training data allows for more effective downstream models training. 
Additionally, ICA enables the use of compact models like BERT for inference, which are significantly more efficient to deploy than prompting large LLMs with lengthy demonstrations \cite{bertsch-etal-2025-context}. 
This makes ICA more suitable for real-world NER applications, particularly in resource-constrained settings.

Finally, we find that applying targeted refinement strategies further improves the quality of ICA. 
Among them, the proposed Error-Aware Refinement (EAR) consistently outperforms other variants across all domains, yielding an average gain of 1.79 F1 points over standard ICA. 
This demonstrates that explicitly addressing common LLM annotation errors, such as missing entities, spurious entities, and type mismatches, can lead to more accurate and reliable supervision for downstream model training.
We further include a case study in Appendix~\ref{apx:case_study} to qualitatively illustrate the effectiveness of these refinement strategies.
We also provide a detailed cost analysis and further discussion on deployment scalability of our ICA approach in Appendix~\ref{apx:cost_study}



\paragraph{Effectiveness of Refinement at Inference Time.}
A natural question is whether our refinement strategies also help when applied directly at inference rather than as a data-annotation step. We apply Self-Correction and EAR to \textsc{ICL-MS} outputs (see Table~\ref{tab:main-results}). 
\textsc{ICL-MS w/ EAR} improves over \textsc{ICL-MS} by +2.41 average F1 (79.44 $\rightarrow$ 81.85), confirming that our refinement strategies are effective beyond the data-annotation setting. We note that prior work has observed limitations of generic intrinsic self-correction in zero-shot reasoning tasks~\citep{huang2023large, hao2025understanding}. 
Our findings, in line with domain-specific studies such as ProgGen~\citep{heng-etal-2024-proggen}, suggest that these limitations can be mitigated by (i) replacing open-ended self-verification with task-specific few-shot demonstrations of error patterns paired with corrections, and (ii) decomposing refinement into focused sub-tasks (spurious, missing, type) rather than asking the model to fix everything at once. Nevertheless, the \textsc{ICA}-based approach still outperforms \textsc{ICL-MS w/ EAR} by 2.95 F1, indicating that converting LLM predictions into downstream training data yields additional gains beyond what refinement alone provides, while also enabling efficient BERT-based deployment.

\section{Additional Studies and Discussion} \label{sec:discussion}
In this section, we report a series of experiments to further understand the behavior and requirements of LLMs when used for in-context annotation in low-resource NER settings. We aim to answer the following three research questions: \textbf{RQ1:} How much LLM-annotated data is needed to train an effective small NER model? \textbf{RQ2:} How much human-annotated data is required for LLMs to generate high-quality annotations? \textbf{RQ3:} How important is the choice of LLM? Can smaller or open-source models perform competitively?
We conduct all studies on the \textit{AI} and \textit{Literature} domains and using ICA with EAR to annotate the data.
In the following content of this section, ``Previous SOTA'' refers to DTrans-MPrompt~\cite{zhang2024cross}, the strongest cross-domain baseline from Table~\ref{tab:main-results}.

\subsection{RQ1: How Much LLM-Annotated Data is Needed?}

\begin{figure}[tb]
   \begin{minipage}{0.24\textwidth}
     \centering
     \includegraphics[width=\linewidth]{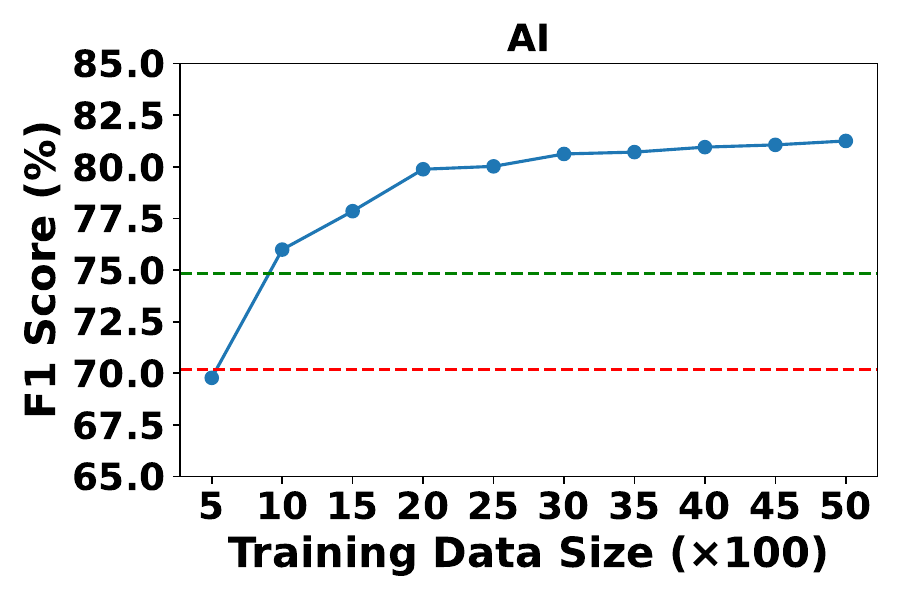}
   \end{minipage}\hfill
   \begin{minipage}{0.24\textwidth}
     \centering
     \includegraphics[width=\linewidth]{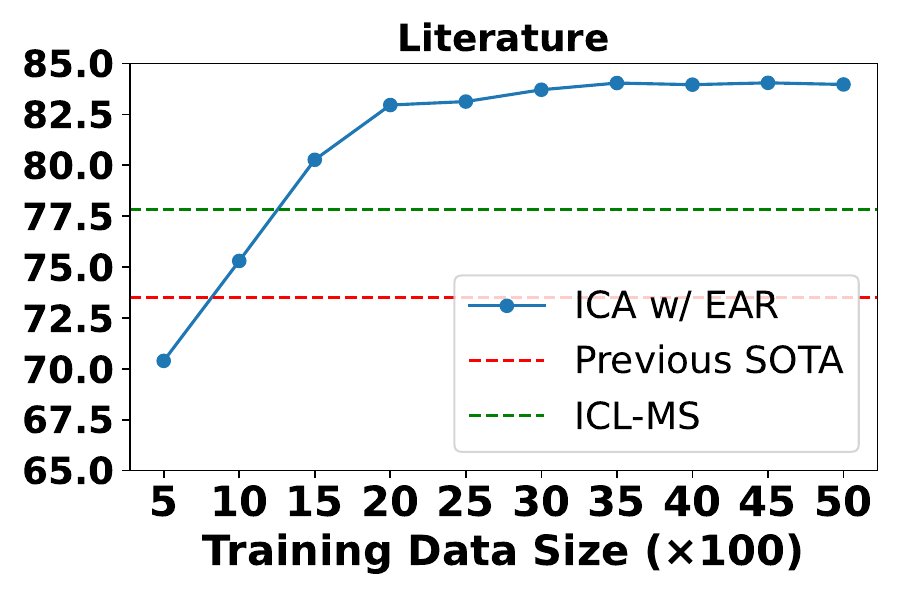}
   \end{minipage}
\vspace{-5pt}
\caption{Performance trends on the AI and Literature domains as the annotated training data increases.}
\label{fig:ner_trends}
\end{figure}

To investigate how much LLM-annotated data is needed to train an effective NER model, we conduct an annotation scale-up experiment. Specifically, we use ICA with EAR to label between 0.5k and 5k unlabeled sentences in increments of 0.5k. 
The resulting data is then used to train a BERT-based NER model, and we evaluate its performance on the test set of the corresponding domain.
As shown in Figure~\ref{fig:ner_trends}, performance improves rapidly with the first 1.5k-2k annotated examples, but plateaus after around 2.5k samples. 
This saturation suggests that early gains come from learning general entity patterns, while additional examples offer diminishing returns due to redundancy or limited diversity in the unlabeled pool. Based on this trend, we find that annotating 1.5k–2.5k sentences is a practical and efficient choice for low-resource settings.
Notably, with just 1k-1.5k annotated examples, our method already surpasses both ICL-MS and previous state-of-the-art methods. This demonstrates the data efficiency of our ICA framework and its ability to produce high-quality training data with minimal human supervision.


\subsection{RQ2: How Much human-annotated Data is Needed for Effective ICA?}

\begin{figure}[tb]
   \begin{minipage}{0.24\textwidth}
     \centering
     \includegraphics[width=\linewidth]{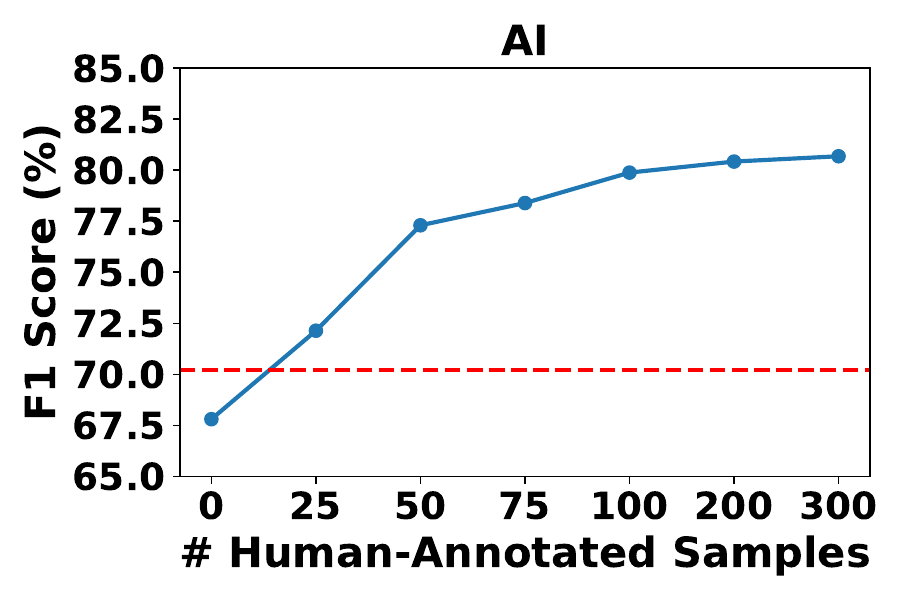}
   \end{minipage}\hfill
   \begin{minipage}{0.24\textwidth}
     \centering
     \includegraphics[width=\linewidth]{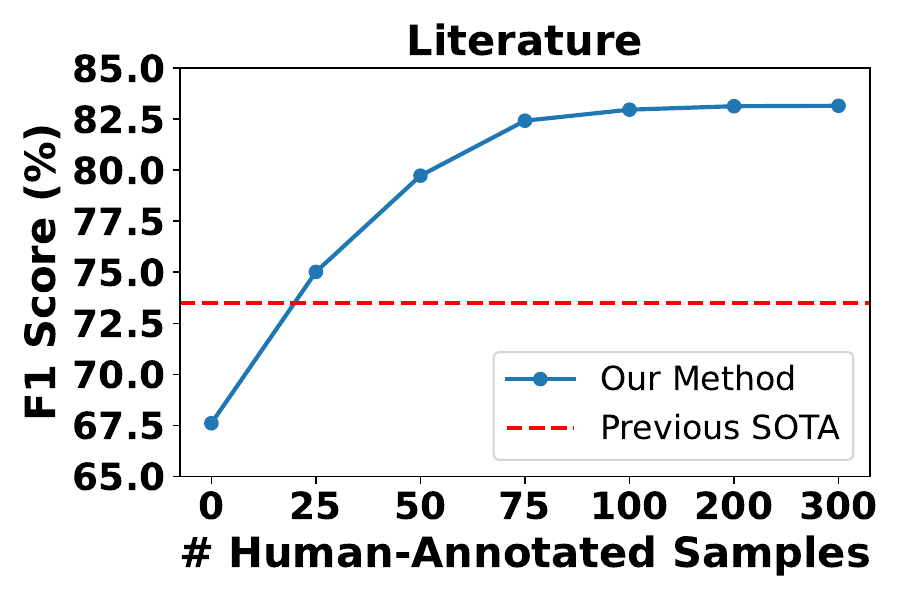}
   \end{minipage}
\vspace{-5pt}
\caption{Performance trends on AI and Literature as the number of human-annotated samples increases.}
\label{fig:human_annotated_study}
\end{figure}

This experiment investigates how much human-annotated data is necessary to guide LLMs for effective in-context annotation. 
Specifically, we vary the number of human-annotated examples settings used in the ICA prompt from 0 to 300. 
We then annotated 2k samples for each setting and used those data to train a BERT-based NER model.

As shown in Figure~\ref{fig:human_annotated_study}, performance improves significantly as the number of human-annotated examples increases from 0 to around 75–100. 
Beyond this point, gains become marginal, with nearly flat curves on both AI and Literature domains. 
These results suggest that using approximately 75–100 human-annotated examples is sufficient to enable strong ICA performance. 
Such a requirement is practically feasible, as collecting this amount of annotated data typically takes only a few hours. 
It strikes an effective balance between human effort and LLM utility for building high-quality NER training sets in low-resource domains.

\subsection{RQ3: How Important is the Choice of LLM?}
\begin{figure}[!t]
   \begin{minipage}{0.24\textwidth}
     \centering
     \includegraphics[width=\linewidth]{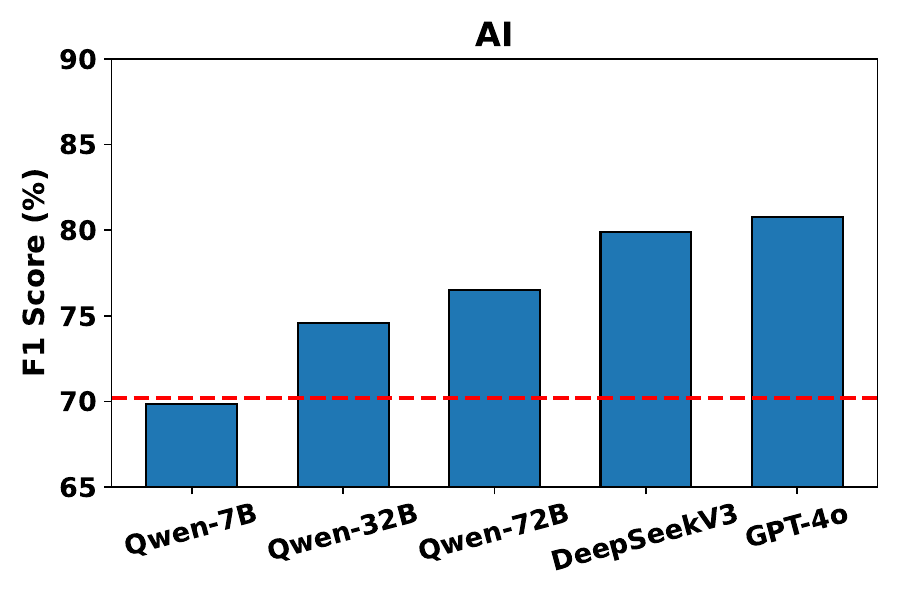}
   \end{minipage}\hfill
   \begin{minipage}{0.24\textwidth}
     \centering
     \includegraphics[width=\linewidth]{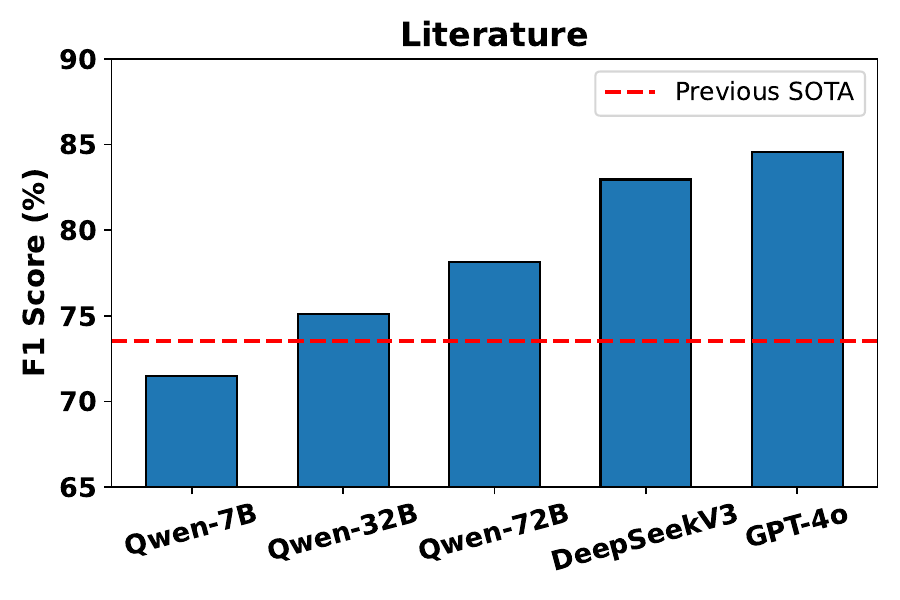}
   \end{minipage}
\vspace{-5pt}
\caption{Comparison of LLMs on AI and Literature. The red dashed line indicates SOTA's performance.}
\label{fig:llm_comparison}
\end{figure}

To study the impact of LLM choice on annotation quality, we conduct an experiment using different LLMs to perform in-context annotation. 
Specifically, each model is used to label 2k sentences for both the AI and Literature domains using our ICA framework with EAR. 
The annotated data is then used to train a BERT-based NER model. 
We compare several models of varying scale and architecture, including the frontier model \textbf{GPT-4o}, and three variants of \textbf{Qwen-2.5} with 7B, 32B, and 72B parameters. DeepSeekV3 is also included as a competitive baseline.
As shown in Figure~\ref{fig:llm_comparison}, the choice of LLM significantly impacts downstream performance. 
Larger and more capable models yield higher-quality annotations that translate into better NER performance. 
For both domains, GPT-4o achieves the best results, followed closely by DeepSeekV3 and Qwen-72B. In contrast, smaller models such as Qwen-7B lag behind by a wide margin, particularly on the more complex Literature domain.
These results suggest that model scale and pretraining quality both matter for in-context annotation. While open-source models can be competitive when scaled up (e.g., Qwen-72B), there remains a notable gap between smaller LLMs and the best proprietary models. This highlights a trade-off between annotation quality and model accessibility.

\section{Related Work}
\paragraph{Named Entity Recognitions.} 
Large language models have been applied to NER via zero-shot and few-shot prompting, but early attempts fell short of the accuracy of task-specific models \cite{jimenez-gutierrez-etal-2022-thinking, ashok2023promptner}
Recent work focuses on closing this gap by better integrating task knowledge into prompts and more advanced prompt techniques \cite{zhang-etal-2025-survey}. 
For example, PromptNER augments few-shot examples with explicit entity type definitions, prompting an LLM to list entities in a text along with explanations;
\citet{xie-etal-2023-empirical} proposes an inference strategy combining problem decomposition, syntactic augmentation, and a two-stage majority voting scheme, significantly improving ChatGPT's zero-shot NER performance.
\citet{xie-etal-2024-self} involves self-annotation of unlabeled corpora, selection of reliable annotations, and in-context learning with retrieved demonstrations.
\citet{heng-etal-2024-proggen} introduces a cost-effective framework that guides large language models to self-reflect on domain-specific attributes and pre-generate entity terms, thereby creating high-quality NER datasets.
Another batch approach is to design general-purpose NER systems inspired by LLMs' flexibility. 
Methods such as GLiNER\cite{zaratiana-etal-2024-gliner}, NuNER\cite{bogdanov-etal-2024-nuner}, GliNER \cite{sainz2023gollie}, and UniversalNER \cite{zhouuniversalner} leverage large-scale synthetic NER data for extensive training, thereby achieving strong zero-shot performance.
Another line of work builds unified NER frameworks to handle flat, nested, and discontinuous entities.
UIE introduces a text-to-structure generation framework that unifies multiple IE tasks via schema-based prompting and a structured extraction language \cite{lu2022unified};
LasUIE extends this by inducing latent syntactic structures and integrating them into the generation process \cite{fei2022lasuie}.
In the scientific domain, several corpora have been developed to support NER over scholarly texts, including DMDD for dataset mention detection \cite{pan-etal-2023-dmdd}, SciDMT for joint detection of datasets, methods, and tasks \cite{pan-etal-2024-scidmt}, and SciER for entity and relation extraction from full-text scientific publications \cite{zhang-etal-2024-scier}.
Since large-scale scientific corpora often rely on weak or distant supervision, label-cleaning approaches such as DynClean \cite{zhang-etal-2025-dynclean} have been proposed to improve annotation quality by leveraging training dynamics to identify mislabeled instances.
However, these methods typically require substantial computational resources for pre-training and synthetic data generation.
In contrast, our framework can serve as a cost-effective annotation generator, producing high-quality labels for training unified NER models.



\paragraph{Many-Shot In-Context Learning.}
Many studies suggested that using a small number of demonstrations, typically less than one hundred, was sufficient to achieve strong performance\cite{highmore2024context, min-etal-2022-rethinking}, and some even indicated that increasing the number of demonstrations could potentially hurt ICL performance \cite{chen-etal-2023-many}.
Recent long-context architectures expand context length by orders of magnitude (from a few thousand tokens in GPT-3 to around 1 million tokens in recent research models) \cite{agarwal2024many}, enabling many-shot in-context learning with hundreds or thousands of examples in a single prompt.
Studies have observed that providing far more demonstrations often yields significant accuracy gains compared to the traditional few-shot regime \cite{agarwal2024many, bertsch-etal-2025-context}.
In fact, as the number of prompt examples grows into the hundreds, models begin to approach or even match fine-tuned performance on certain tasks \cite{agarwal2024many}, indicating that sufficiently large prompts allow LLMs to overcome some pre-training biases.
One recent 
work \citet{song-etal-2025-many} study whether many-shot ICL can improve the reliability of LLMs as evaluators, and finds that using many-shot prompts—especially with reference answers—leads to more consistent and accurate evaluation results.
These findings demonstrate the feasibility of leveraging many-shot ICL to further enhance the performance of LLM-based ICL and to address the challenge of limited training data.




\section{Conclusion}

This work provides the first comprehensive investigation of many-shot in-context learning (ICL) for Named Entity Recognition (NER), demonstrating its viability as both an inference-time strategy and a scalable annotation framework. 
We illustrate that with only a modest number of seed annotations (\textit{around a hundred}), many-shot in-context annotation can generate high-quality labeled data, enabling significant downstream performance gains.

Building on these findings, we propose the ICA framework, which uses LLMs as offline annotators to efficiently generate and refine labeled data with minimal human effort.
By decoupling LLM annotation from deployment, ICA enables compact models like BERT to be trained on high-quality data and deployed cost-effectively at scale.
In low-resource settings, ICA achieves approximately 10\% absolute F1 improvement over previous SOTA methods.
Our findings highlight the promise of LLM-based ICL and annotation frameworks like ICA in improving data efficiency and reducing annotation costs for NER and related tasks.

\section*{Limitations}

While our study demonstrates the promise of many-shot in-context learning (ICL) for NER, several limitations remain. While our method reduces the need for large-scale manual annotation, it still requires a moderate number of high-quality seed labels and introduces computational overhead from repeated LLM inference. In addition, ICL offers little transparency into why certain outputs are generated or how LLMs adapt to increasing demonstration size, complicating efforts to debug or audit model behavior in sensitive applications. Finally, our experiments focus on English-language NER in a limited set of domains. Further evaluation is needed to understand the generalizability of many-shot ICL to other languages, domains, and entity types, especially those with more complex or ambiguous annotation schemes. Future work can address these limitations by exploring hybrid models that combine the efficiency of ICL with targeted fine-tuning, extending evaluations to multilingual and multi-domain settings.

\section*{Acknowledgements}
This work was supported by the National Science Foundation awards III-2107213, III-2107518, 2026513, and ITE-2333789.
We thank the reviewers and the area chair for their constructive feedback and suggestions, which have substantially improved the quality of this paper.

\bibliography{custom,anthology}

\clearpage

\appendix

\section{Additional Details of Many-Shot ICL Study}

\subsection{Dataset Statistics} \label{apx:micl_dataset}

We consider 4 NER datasets from different domains: CoNLL2003 \cite{tjong-kim-sang-de-meulder-2003-introduction}, MIT-Movie and MIT-Restaurant \cite{mitner}, WNUT 2017 \cite{derczynski-etal-2017-results}.

\begin{table}[h]
    \centering
    \resizebox{0.468\textwidth}{!}{
    \begin{tabular}{c|c|c|c}
        \hline
        \textbf{Name} & \textbf{Domain} & \begin{tabular}[c]{@{}c@{}}\textbf{Entity} \\\textbf{Types}\end{tabular} 
  & \textbf{Data size} \\
        \hline
        \begin{tabular}[c]{@{}c@{}}CoNLL\\ 2003\end{tabular}      & News             & 4  & 14,041 / 3,250 / 3,453 \\ \hline
        \begin{tabular}[c]{@{}c@{}}WNUT\\ 2017\end{tabular}        & SNS              & 6  & 1,000 / 1,008 / 1,287 \\ \hline
        \begin{tabular}[c]{@{}c@{}}MIT\\ Restaurant\end{tabular}   & \begin{tabular}[c]{@{}c@{}}Restaurant\\review \end{tabular}   & 8  & 7,660 / - / 1,521 \\ \hline
        \begin{tabular}[c]{@{}c@{}}MIT\\ Movie\end{tabular}      &  \begin{tabular}[c]{@{}c@{}}Movie\\review \end{tabular}      & 12 & 7,816 / - / 1,953 \\
        \hline
    \end{tabular}
    }
    \caption{Four NER datasets are considered. Data size is the number of sentences in training/validation/test set.}
\end{table}

\subsection{Many-shot ICL Models and Settings Details} \label{apx:micl_exp_details}

\paragraph{LLMs details.}

The majority of the analysis in the paper concerns these five LLMs models:
    
    
    

\begin{itemize}
    \item \textbf{GPT-4o} \cite{gpt4o} is a recent multimodal model developed by OpenAI, designed for fast and cost-efficient inference. It supports long-context inputs and demonstrates strong reasoning capabilities.

    \item \textbf{DeepSeekV3} \cite{deepseekv3} is a decoder-only language model optimized for both efficiency and long-context understanding.

    \item \textbf{Qwen2.5-7B and Qwen2.5-72B} \cite{qwen} are multilingual decoder-only models with trained context lengths of 128k and 131k tokens, respectively. We use their instruction-tuned variants in our experiments.

    \item \textbf{LLaMA3.1-8B and LLaMA3.1-70B} \cite{llama} are the latest models in Meta’s LLaMA series, featuring improvements in both training data and model architecture. Both variants support a trained context length of 128k tokens. We employ the instruction-tuned versions in all evaluations.
\end{itemize}

We use the vLLM\footnote{https://docs.vllm.ai/en/stable/} v0.8.5 to deploy the Qwen2.5 and LlaMa 3.1 LLMs on a machine with 4 A100 GPUs. For the DeepSeekV3 and GPT-4o, we use the official provided API. All the experiments results are obained during the January 20, 2025 to May 10, 2025.

\paragraph{BERT Fine-Tuning detals.}
All BERT-FT models in this paper refer to the fine-tuned \textit{bert-base-cased}\footnote{https://huggingface.co/bert-base-cased} version. 
We use the same combnition of hyperparameters for all datasets: the learning rate is set as 1e-5; the training batch size is 16, all results are based on the fine-tuned results of 5 epochs and report average of five runs, each with a different random seed.
Experiments were conducted using a single A100 GPU card with PyTorch 2.7.0.
The average running time on the CoNLL03 dataset is 34 seconds/epoch.

\subsection{Baseline Implementation Details}
\label{apx:baselines}

\paragraph{ProgGen.}
We implemented ProgGen~\cite{heng-etal-2024-proggen} on CrossNER by following the authors' released code and recipe for data generation and training. We adopted the best-performing configuration reported in the paper—combining their data generation and refinement pipeline with the strongest downstream model setup—and ran all experiments under this setting to ensure comparability with their reported results.

\paragraph{GliNER.}
We fine-tuned the GliNER encoder~\cite{zaratiana-etal-2024-gliner} on CrossNER using 100 human-annotated examples per domain to match the data budget of our ICA framework.

\paragraph{Other Baselines.}
For the remaining baselines, we report results on CrossNER directly from the original sources for fair comparison. Note that many of these baselines use not only the 100 human-annotated training examples but also additional validation data for hyperparameter tuning.

\subsection{Detail Many-shot ICL for NER results} \label{apx:micl_ner_detail}

\begin{figure*}[t]
  \centering
  \begin{minipage}[t]{0.495\textwidth}
    \centering
    \includegraphics[width=0.98\linewidth]{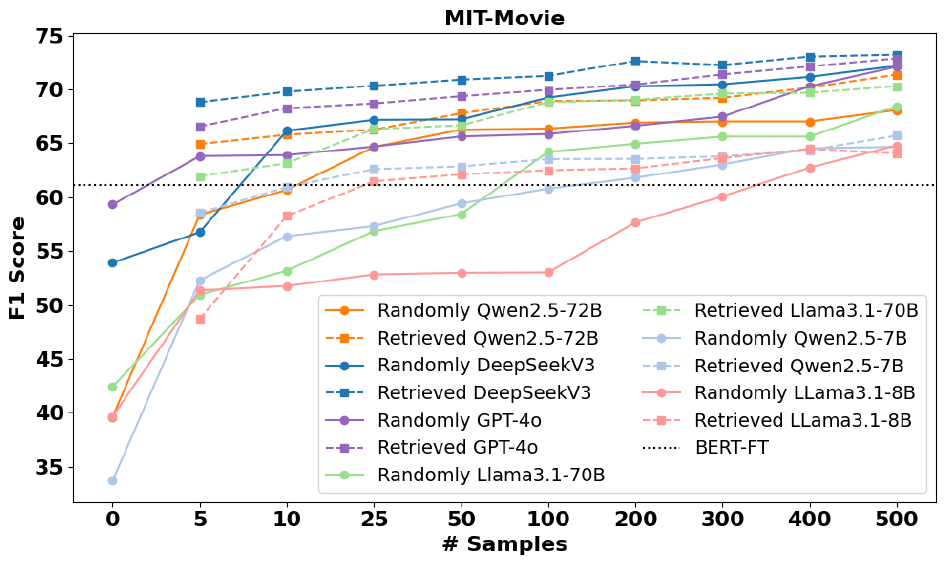}
    \vspace{-8pt}
  \end{minipage}
  \hfill
  \begin{minipage}[t]{0.495\textwidth}
    \centering
    \includegraphics[width=0.98\linewidth]{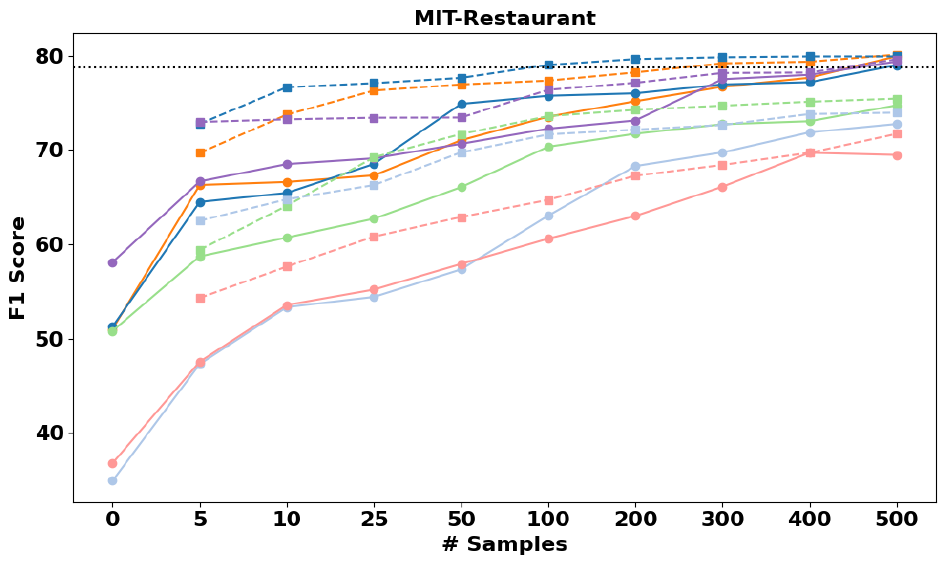}
    \vspace{-8pt}
  \end{minipage}
  \vspace{-16pt}
  \caption{Complete performance results on the MIT-Movie and MIT-Restaurant datasets as the number of samples increases.}
  \label{fig:many-shots-trends-complete1}
\end{figure*}

\begin{figure*}[t]
  \centering
  \begin{minipage}[t]{0.495\textwidth}
    \centering
    \includegraphics[width=\linewidth]{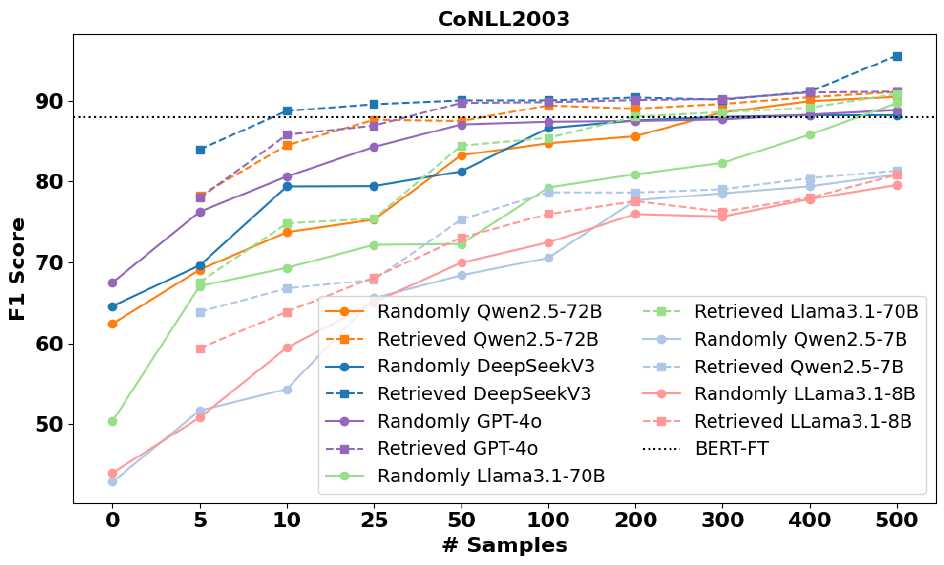}
    \vspace{-8pt}
  \end{minipage}
  \hfill
  \begin{minipage}[t]{0.495\textwidth}
    \centering
    \includegraphics[width=0.98\linewidth]{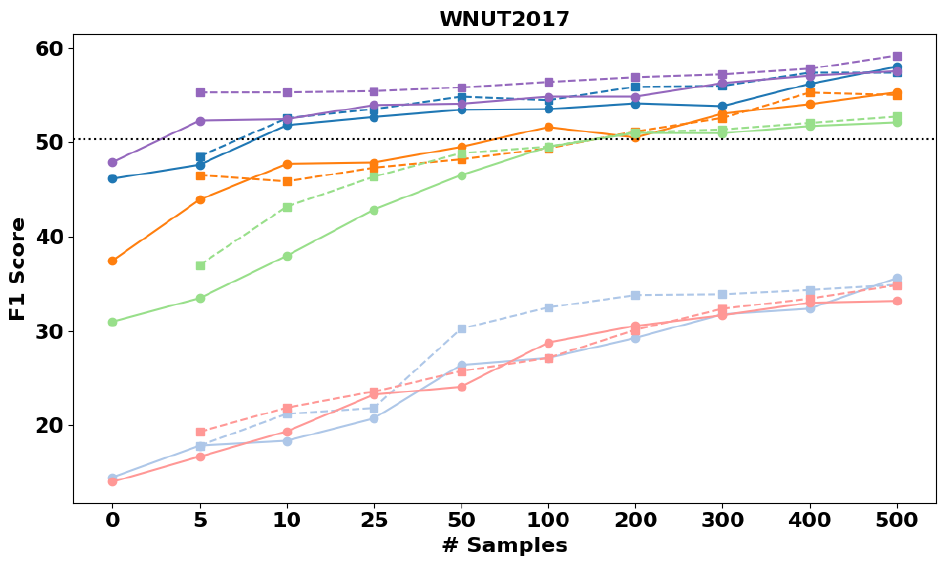}
    \vspace{-8pt}
  \end{minipage}
  \vspace{-16pt}
  \caption{Complete performance results on the CoNLL2003 and WNUT2017 datasets as the number of samples increases.}
  \label{fig:many-shots-trends-complete2}
\end{figure*}

Table~\ref{tab:large-table} provides the complete performance results of all seven language models under both random and retrieval-based ICL settings across the four datasets.
Figure~\ref{fig:many-shots-trends-complete1} and~\ref{fig:many-shots-trends-complete2} presents the performance trends of all models as the number of in-context examples increases.

From the figures, we observe a consistent trend: the performance of LLMs generally improves as the number of in-context examples increases, especially under the random sampling setting.
For smaller models such as Qwen2.5-7B and LLaMA3.1-8B, although their final performance may not match that of the BERT-FT baseline, they still exhibit a clear many-shot ICL improvement trend as the number of demonstrations grows.


\begin{table*}[]
\centering
\scalebox{0.42}{
\begin{tabular}{ll|llll|lll|lll|lll}
\hline
 &
  \textbf{} &
  \textbf{} &
  \multicolumn{3}{l|}{\textbf{MIT-Movie}} &
  \multicolumn{3}{l|}{\textbf{MIT-Res}} &
  \multicolumn{3}{l|}{\textbf{CoNLL2003}} &
  \multicolumn{3}{l}{\textbf{WNUT2017}} \\ \hline
 &
  \textbf{} &
  \textbf{\# Shots} &
  \textbf{P} &
  \textbf{R} &
  \textbf{F1} &
  \textbf{P} &
  \textbf{R} &
  \textbf{F1} &
  \textbf{P} &
  \textbf{R} &
  \textbf{F1} &
  \textbf{P} &
  \textbf{R} &
  \textbf{F1} \\ \hline
\multicolumn{2}{l|}{\textbf{SOTA}}                               &     & -     & -     & 71.20 & -     & -     & 79.60 & -     & -     & 94.30 & -     & -     & 52.80 \\ \hline
\multicolumn{2}{l|}{\textbf{BERT}}                               &     & 58.92 & 63.80 & 61.17 & 76.33 & 81.43 & 78.77 & 86.79 & 89.12 & 87.92 & -     & -     & 50.30 \\ \hline
\multicolumn{2}{l|}{\multirow{10}{*}{\textbf{Qwen-72B}}}         & 0   & 46.20 & 34.70 & 39.63 & 56.94 & 46.19 & 51.01 & 64.37 & 60.61 & 62.43 & 30.16 & 49.44 & 37.46 \\
\multicolumn{2}{l|}{}                                            & 5   & 56.18 & 60.73 & 58.37 & 64.77 & 67.88 & 66.29 & 65.13 & 73.59 & 69.11 & 33.87 & 62.45 & 43.92 \\
\multicolumn{2}{l|}{}                                            & 10  & 58.62 & 62.90 & 60.68 & 74.96 & 79.71 & 66.61 & 71.08 & 76.62 & 73.75 & 44.03 & 52.04 & 47.70 \\
\multicolumn{2}{l|}{}                                            & 25  & 63.28 & 66.10 & 64.66 & 65.52 & 69.21 & 67.31 & 77.10 & 73.59 & 75.30 & 46.34 & 49.44 & 47.84 \\
\multicolumn{2}{l|}{}                                            & 50  & 63.21 & 69.63 & 66.27 & 68.73 & 73.51 & 71.04 & 80.63 & 86.09 & 83.27 & 44.51 & 55.76 & 49.50 \\
\multicolumn{2}{l|}{}                                            & 100 & 64.28 & 68.61 & 66.37 & 72.13 & 75.00 & 73.54 & 82.99 & 86.58 & 84.75 & 47.50 & 56.51 & 51.61 \\
\multicolumn{2}{l|}{}                                            & 200 & 64.71 & 69.29 & 66.92 & 74.39 & 75.99 & 75.18 & 84.89 & 86.36 & 85.62 & 45.65 & 56.51 & 50.50 \\
\multicolumn{2}{l|}{}                                            & 300 & 63.84 & 70.55 & 67.03 & 75.20 & 78.31 & 76.72 & 86.57 & 90.69 & 88.58 & 48.32 & 58.74 & 53.02 \\
\multicolumn{2}{l|}{}                                            & 400 & 64.68 & 69.57 & 67.03 & 74.66 & 80.96 & 77.68 & 87.99 & 91.99 & 89.95 & 54.55 & 53.53 & 54.03 \\
\multicolumn{2}{l|}{}                                            & 500 & 65.51 & 70.89 & 68.09 & 77.98 & 82.03 & 79.96 & 88.89 & 92.17 & 90.50 & 46.59 & 67.96 & 55.28 \\ \hline
\multicolumn{2}{l|}{\multirow{9}{*}{\textbf{Qwen-72B (BM25)}}}   & 5   & 62.89 & 67.12 & 64.94 & 67.23 & 72.35 & 69.70 & 76.16 & 80.51 & 78.27 & 37.73 & 60.59 & 46.50 \\
\multicolumn{2}{l|}{}                                            & 10  & 62.64 & 69.29 & 65.80 & 71.36 & 76.32 & 73.76 & 82.92 & 86.15 & 84.50 & 39.67 & 54.28 & 45.84 \\
\multicolumn{2}{l|}{}                                            & 25  & 63.49 & 69.29 & 66.27 & 73.39 & 79.47 & 76.31 & 86.19 & 89.18 & 87.66 & 43.34 & 52.04 & 47.30 \\
\multicolumn{2}{l|}{}                                            & 50  & 66.78 & 68.84 & 67.79 & 73.75 & 80.46 & 76.96 & 85.68 & 89.39 & 87.50 & 43.32 & 54.28 & 48.18 \\
\multicolumn{2}{l|}{}                                            & 100 & 67.88 & 69.98 & 68.92 & 74.09 & 80.96 & 77.37 & 88.21 & 90.69 & 89.43 & 48.23 & 50.56 & 49.36 \\
\multicolumn{2}{l|}{}                                            & 200 & 67.58 & 70.43 & 68.98 & 76.24 & 80.35 & 78.24 & 87.63 & 90.48 & 89.03 & 48.34 & 54.28 & 51.14 \\
\multicolumn{2}{l|}{}                                            & 300 & 67.54 & 71.00 & 69.23 & 76.07 & 82.62 & 79.21 & 87.76 & 91.56 & 89.62 & 52.00 & 53.16 & 52.57 \\
\multicolumn{2}{l|}{}                                            & 400 & 68.10 & 72.37 & 70.17 & 77.69 & 81.14 & 79.38 & 89.11 & 91.84 & 90.45 & 57.96 & 52.79 & 55.25 \\
\multicolumn{2}{l|}{}                                            & 500 & 71.07 & 71.67 & 71.37 & 77.64 & 82.78 & 80.13 & 89.95 & 92.27 & 91.10 & 50.23 & 60.77 & 55.00 \\ \hline
\multicolumn{2}{l|}{\multirow{10}{*}{\textbf{DeepSeekV3}}}       & 0   & 51.67 & 56.39 & 53.93 & 56.98 & 46.59 & 51.26 & 61.40 & 68.18 & 64.62 & 39.52 & 55.39 & 46.13 \\
\multicolumn{2}{l|}{}                                            & 5   & 58.33 & 55.26 & 56.76 & 64.80 & 64.18 & 64.49 & 64.46 & 75.76 & 69.65 & 39.05 & 60.97 & 47.61 \\
\multicolumn{2}{l|}{}                                            & 10  & 65.15 & 67.24 & 66.18 & 65.45 & 65.45 & 65.45 & 76.77 & 82.25 & 79.41 & 53.13 & 50.56 & 51.81 \\
\multicolumn{2}{l|}{}                                            & 25  & 64.79 & 69.75 & 67.18 & 66.77 & 70.36 & 68.52 & 77.41 & 81.60 & 79.45 & 52.19 & 53.16 & 52.67 \\
\multicolumn{2}{l|}{}                                            & 50  & 64.76 & 69.86 & 67.22 & 73.04 & 76.86 & 74.90 & 79.50 & 83.12 & 81.27 & 53.33 & 53.53 & 53.43 \\
\multicolumn{2}{l|}{}                                            & 100 & 67.17 & 71.46 & 69.25 & 74.09 & 77.50 & 75.76 & 84.39 & 88.96 & 86.62 & 56.38 & 50.93 & 53.52 \\
\multicolumn{2}{l|}{}                                            & 200 & 68.32 & 72.37 & 70.29 & 74.02 & 78.13 & 76.02 & 87.47 & 87.66 & 87.57 & 52.25 & 56.13 & 54.12 \\
\multicolumn{2}{l|}{}                                            & 300 & 67.97 & 73.17 & 70.48 & 74.66 & 79.40 & 76.96 & 87.08 & 88.96 & 88.01 & 60.75 & 48.33 & 53.83 \\
\multicolumn{2}{l|}{}                                            & 400 & 68.57 & 73.97 & 71.17 & 75.84 & 78.61 & 77.20 & 87.29 & 89.18 & 88.22 & 60.52 & 52.42 & 56.18 \\
\multicolumn{2}{l|}{}                                            & 500 & 70.39 & 74.09 & 72.19 & 76.44 & 81.78 & 79.02 & 87.29 & 89.18 & 88.22 & 56.91 & 59.12 & 57.99 \\ \hline
\multicolumn{2}{l|}{\multirow{9}{*}{\textbf{DeepSeekV3 (BM25)}}} & 5   & 66.17 & 71.69 & 68.82 & 69.77 & 76.07 & 72.78 & 80.80 & 87.45 & 83.99 & 41.41 & 58.56 & 48.51 \\
\multicolumn{2}{l|}{}                                            & 10  & 67.38 & 72.37 & 69.79 & 74.11 & 79.40 & 76.66 & 87.90 & 89.61 & 88.75 & 51.04 & 54.14 & 52.55 \\
\multicolumn{2}{l|}{}                                            & 25  & 67.91 & 72.95 & 70.34 & 75.08 & 79.24 & 77.10 & 88.91 & 90.26 & 89.58 & 54.65 & 52.42 & 53.51 \\
\multicolumn{2}{l|}{}                                            & 50  & 67.71 & 74.43 & 70.91 & 77.36 & 77.97 & 77.66 & 89.01 & 91.13 & 90.05 & 57.03 & 52.79 & 54.83 \\
\multicolumn{2}{l|}{}                                            & 100 & 68.34 & 74.43 & 71.26 & 76.91 & 81.30 & 79.04 & 89.01 & 91.13 & 90.05 & 48.29 & 62.43 & 54.46 \\
\multicolumn{2}{l|}{}                                            & 200 & 69.63 & 75.91 & 72.64 & 77.79 & 81.62 & 79.66 & 89.08 & 91.77 & 90.41 & 56.50 & 55.25 & 55.87 \\
\multicolumn{2}{l|}{}                                            & 300 & 68.87 & 76.03 & 72.27 & 77.58 & 82.25 & 79.85 & 89.19 & 91.13 & 90.15 & 58.23 & 53.90 & 55.98 \\
\multicolumn{2}{l|}{}                                            & 400 & 69.82 & 76.60 & 73.05 & 77.89 & 82.09 & 79.94 & 90.06 & 92.21 & 91.12 & 54.77 & 60.22 & 57.37 \\
\multicolumn{2}{l|}{}                                            & 500 & 70.35 & 76.37 & 73.23 & 77.61 & 82.41 & 79.94 & 90.59 & 91.50 & 91.04 & 55.73 & 59.12 & 57.37 \\ \hline
\multicolumn{2}{l|}{\multirow{10}{*}{\textbf{GPT-4o}}}           & 0   & 58.18 & 60.50 & 59.32 & 63.71 & 53.41 & 58.10 & 62.71 & 73.16 & 67.53 & 39.42 & 60.97 & 47.88 \\
\multicolumn{2}{l|}{}                                            & 5   & 62.10 & 65.64 & 63.82 & 64.82 & 68.62 & 66.67 & 77.04 & 75.54 & 76.28 & 45.58 & 61.34 & 52.30 \\
\multicolumn{2}{l|}{}                                            & 10  & 63.12 & 64.72 & 63.91 & 66.62 & 70.52 & 68.51 & 79.70 & 81.60 & 80.64 & 48.14 & 57.62 & 52.45 \\
\multicolumn{2}{l|}{}                                            & 25  & 63.15 & 66.32 & 64.70 & 68.37 & 69.89 & 69.12 & 83.76 & 84.85 & 84.30 & 50.99 & 57.25 & 53.94 \\
\multicolumn{2}{l|}{}                                            & 50  & 63.31 & 68.15 & 65.64 & 69.42 & 71.95 & 70.66 & 87.36 & 86.80 & 87.08 & 47.47 & 62.83 & 54.08 \\
\multicolumn{2}{l|}{}                                            & 100 & 63.14 & 68.84 & 65.87 & 70.31 & 74.33 & 72.27 & 86.94 & 87.88 & 87.41 & 51.13 & 59.11 & 54.83 \\
\multicolumn{2}{l|}{}                                            & 200 & 63.59 & 69.98 & 66.63 & 71.69 & 74.64 & 73.14 & 86.32 & 88.74 & 87.51 & 52.00 & 57.99 & 54.83 \\
\multicolumn{2}{l|}{}                                            & 300 & 65.34 & 69.76 & 67.48 & 74.45 & 80.86 & 77.52 & 86.68 & 88.74 & 87.70 & 54.33 & 58.36 & 56.27 \\
\multicolumn{2}{l|}{}                                            & 400 & 68.32 & 72.37 & 70.29 & 74.51 & 81.82 & 77.99 & 86.82 & 89.83 & 88.30 & 53.59 & 60.97 & 57.04 \\
\multicolumn{2}{l|}{}                                            & 500 & 69.84 & 74.54 & 72.11 & 76.50 & 82.54 & 79.40 & 88.94 & 88.74 & 88.84 & 53.33 & 62.45 & 57.53 \\ \hline
\multicolumn{2}{l|}{\multirow{9}{*}{\textbf{GPT-4o (BM25)}}}     & 5   & 64.83 & 68.38 & 66.56 & 71.54 & 74.48 & 72.98 & 68.29 & 90.91 & 77.99 & 54.09 & 56.51 & 55.27 \\
\multicolumn{2}{l|}{}                                            & 10  & 66.41 & 70.21 & 68.26 & 70.92 & 75.75 & 73.26 & 83.44 & 88.31 & 85.80 & 52.33 & 58.58 & 55.28 \\
\multicolumn{2}{l|}{}                                            & 25  & 66.74 & 70.78 & 68.70 & 71.24 & 75.75 & 73.43 & 86.02 & 87.88 & 86.94 & 49.56 & 62.83 & 55.41 \\
\multicolumn{2}{l|}{}                                            & 50  & 67.49 & 71.35 & 69.37 & 71.01 & 76.07 & 73.45 & 87.78 & 91.77 & 89.74 & 52.27 & 59.85 & 55.81 \\
\multicolumn{2}{l|}{}                                            & 100 & 68.72 & 71.23 & 69.96 & 73.66 & 79.33 & 76.39 & 87.97 & 91.77 & 89.83 & 52.27 & 61.22 & 56.39 \\
\multicolumn{2}{l|}{}                                            & 200 & 68.85 & 72.15 & 70.46 & 75.29 & 79.09 & 77.14 & 89.70 & 90.48 & 90.09 & 55.96 & 57.84 & 56.88 \\
\multicolumn{2}{l|}{}                                            & 300 & 69.79 & 73.06 & 71.39 & 75.96 & 80.53 & 78.18 & 88.54 & 91.99 & 90.23 & 54.15 & 60.59 & 57.19 \\
\multicolumn{2}{l|}{}                                            & 400 & 70.19 & 74.20 & 72.14 & 75.85 & 80.77 & 78.23 & 90.73 & 91.32 & 91.03 & 54.05 & 62.08 & 57.79 \\
\multicolumn{2}{l|}{}                                            & 500 & 70.57 & 75.34 & 72.88 & 77.63 & 81.73 & 79.63 & 89.89 & 92.42 & 91.14 & 57.25 & 61.24 & 59.18 \\ \hline
\multicolumn{2}{l|}{\multirow{10}{*}{\textbf{Llama3.1-70B}}}     & 0   & 43.75 & 41.18 & 42.42 & 43.37 & 61.30 & 50.80 & 40.51 & 66.67 & 50.39 & 26.36 & 37.57 & 30.98 \\
\multicolumn{2}{l|}{}                                            & 5   & 46.09 & 56.94 & 50.94 & 54.79 & 63.22 & 58.71 & 66.24 & 67.97 & 67.09 & 26.17 & 46.41 & 33.47 \\
\multicolumn{2}{l|}{}                                            & 10  & 49.02 & 58.14 & 53.19 & 56.67 & 65.38 & 60.71 & 59.02 & 84.03 & 69.34 & 30.72 & 49.72 & 37.97 \\
\multicolumn{2}{l|}{}                                            & 25  & 52.73 & 61.70 & 56.86 & 58.73 & 67.27 & 62.71 & 60.61 & 89.24 & 72.19 & 38.30 & 48.65 & 42.86 \\
\multicolumn{2}{l|}{}                                            & 50  & 54.03 & 63.62 & 58.43 & 56.91 & 59.12 & 66.06 & 65.44 & 80.74 & 72.29 & 41.67 & 52.63 & 46.51 \\
\multicolumn{2}{l|}{}                                            & 100 & 62.75 & 65.73 & 64.21 & 70.26 & 70.43 & 70.35 & 77.87 & 80.74 & 79.28 & 43.84 & 56.88 & 49.51 \\
\multicolumn{2}{l|}{}                                            & 200 & 63.08 & 66.96 & 64.96 & 70.90 & 72.58 & 71.73 & 76.31 & 86.11 & 80.91 & 46.43 & 56.52 & 50.98 \\
\multicolumn{2}{l|}{}                                            & 300 & 64.68 & 66.61 & 65.63 & 70.93 & 74.64 & 72.74 & 79.17 & 85.76 & 82.33 & 48.03 & 54.28 & 50.96 \\
\multicolumn{2}{l|}{}                                            & 400 & 64.52 & 66.78 & 65.63 & 70.83 & 75.44 & 73.06 & 83.55 & 88.19 & 85.81 & 50.00 & 53.53 & 51.71 \\
\multicolumn{2}{l|}{}                                            & 500 & 67.35 & 69.60 & 68.45 & 72.47 & 77.18 & 74.75 & 87.21 & 92.36 & 89.71 & 49.17 & 55.39 & 52.10 \\ \hline
\multicolumn{2}{l|}{\multirow{9}{*}{\textbf{Llama3.1-70B (BM25)}}} &
  5 &
  58.18 &
  66.26 &
  61.96 &
  48.99 &
  75.48 &
  59.41 &
  55.31 &
  86.81 &
  67.57 &
  31.52 &
  44.75 &
  36.99 \\
\multicolumn{2}{l|}{}                                            & 10  & 61.22 & 65.22 & 63.16 & 54.56 & 77.64 & 64.09 & 65.79 & 86.81 & 74.85 & 38.78 & 48.72 & 43.18 \\
\multicolumn{2}{l|}{}                                            & 25  & 63.95 & 68.89 & 66.33 & 66.67 & 72.00 & 69.23 & 65.45 & 88.96 & 75.41 & 42.22 & 51.35 & 46.34 \\
\multicolumn{2}{l|}{}                                            & 50  & 66.00 & 67.35 & 66.67 & 69.09 & 74.51 & 71.70 & 79.88 & 89.58 & 84.45 & 44.68 & 53.85 & 48.84 \\
\multicolumn{2}{l|}{}                                            & 100 & 66.45 & 71.35 & 68.81 & 71.59 & 75.72 & 73.60 & 81.78 & 89.39 & 85.42 & 42.82 & 58.74 & 49.53 \\
\multicolumn{2}{l|}{}                                            & 200 & 66.72 & 71.53 & 69.04 & 71.85 & 76.86 & 74.27 & 84.39 & 92.01 & 88.04 & 52.34 & 49.81 & 51.05 \\
\multicolumn{2}{l|}{}                                            & 300 & 67.49 & 71.88 & 69.62 & 72.54 & 77.02 & 74.71 & 84.54 & 93.06 & 88.60 & 50.00 & 52.79 & 51.36 \\
\multicolumn{2}{l|}{}                                            & 400 & 66.88 & 72.76 & 69.70 & 72.63 & 77.81 & 75.13 & 85.85 & 92.71 & 89.15 & 52.04 & 52.04 & 52.04 \\
\multicolumn{2}{l|}{}                                            & 500 & 67.26 & 73.64 & 70.30 & 72.69 & 78.45 & 75.46 & 89.33 & 92.42 & 90.85 & 53.01 & 52.42 & 52.71 \\ \hline
\multicolumn{2}{l|}{\multirow{10}{*}{\textbf{Qwen2.5-7B}}}       & 0   & 35.40 & 32.16 & 33.70 & 46.59 & 27.88 & 34.89 & 47.28 & 39.24 & 42.88 & 57.69 & 8.29  & 14.49 \\
\multicolumn{2}{l|}{}                                            & 5   & 51.72 & 52.90 & 52.30 & 46.51 & 48.08 & 47.28 & 66.30 & 42.36 & 51.69 & 59.38 & 10.50 & 17.84 \\
\multicolumn{2}{l|}{}                                            & 10  & 63.38 & 50.79 & 56.39 & 54.52 & 52.16 & 53.32 & 59.50 & 50.00 & 54.34 & 54.05 & 11.05 & 18.35 \\
\multicolumn{2}{l|}{}                                            & 25  & 60.27 & 54.66 & 57.33 & 57.14 & 51.92 & 54.41 & 78.64 & 56.25 & 65.59 & 37.14 & 14.36 & 20.72 \\
\multicolumn{2}{l|}{}                                            & 50  & 60.73 & 58.17 & 59.43 & 62.93 & 52.64 & 57.33 & 82.04 & 58.68 & 68.42 & 35.51 & 20.99 & 26.39 \\
\multicolumn{2}{l|}{}                                            & 100 & 64.31 & 57.65 & 60.80 & 65.13 & 61.06 & 63.03 & 69.97 & 71.18 & 70.57 & 31.62 & 23.76 & 27.13 \\
\multicolumn{2}{l|}{}                                            & 200 & 66.87 & 57.47 & 61.81 & 66.29 & 70.43 & 68.30 & 87.07 & 70.14 & 77.69 & 51.39 & 20.44 & 29.25 \\
\multicolumn{2}{l|}{}                                            & 300 & 62.98 & 63.09 & 63.04 & 68.85 & 70.67 & 69.75 & 81.65 & 75.69 & 78.56 & 28.83 & 35.36 & 31.76 \\
\multicolumn{2}{l|}{}                                            & 400 & 64.95 & 64.15 & 64.54 & 75.60 & 68.51 & 71.88 & 83.52 & 75.69 & 79.42 & 48.35 & 24.31 & 32.35 \\
\multicolumn{2}{l|}{}                                            & 500 & 64.84 & 64.50 & 64.67 & 72.00 & 73.56 & 72.77 & 84.21 & 77.78 & 80.87 & 76.36 & 23.20 & 35.59 \\ \hline
\multicolumn{2}{l|}{\multirow{9}{*}{\textbf{Qwen2.5-7B (BM25)}}} & 5   & 56.77 & 60.46 & 58.55 & 59.44 & 65.87 & 62.49 & 59.17 & 69.44 & 63.90 & 59.38 & 10.50 & 17.84 \\
\multicolumn{2}{l|}{}                                            & 10  & 60.21 & 61.69 & 60.94 & 69.92 & 60.34 & 64.77 & 76.92 & 59.03 & 66.80 & 63.89 & 12.71 & 21.20 \\
\multicolumn{2}{l|}{}                                            & 25  & 64.73 & 60.63 & 62.61 & 63.98 & 68.75 & 66.28 & 79.17 & 59.38 & 67.86 & 52.08 & 13.81 & 21.83 \\
\multicolumn{2}{l|}{}                                            & 50  & 62.81 & 62.92 & 62.86 & 68.85 & 70.67 & 69.75 & 78.28 & 72.57 & 75.32 & 40.00 & 24.31 & 30.24 \\
\multicolumn{2}{l|}{}                                            & 100 & 66.54 & 60.81 & 63.54 & 74.55 & 68.99 & 71.66 & 83.59 & 74.31 & 78.68 & 38.35 & 28.18 & 32.48 \\
\multicolumn{2}{l|}{}                                            & 200 & 63.67 & 63.44 & 63.56 & 70.83 & 73.56 & 72.17 & 84.46 & 73.61 & 78.66 & 33.33 & 34.25 & 33.79 \\
\multicolumn{2}{l|}{}                                            & 300 & 66.48 & 61.34 & 63.80 & 75.58 & 69.95 & 72.66 & 82.33 & 76.04 & 79.06 & 62.69 & 23.20 & 33.87 \\
\multicolumn{2}{l|}{}                                            & 400 & 66.48 & 62.39 & 64.37 & 74.09 & 73.56 & 73.82 & 83.27 & 77.78 & 80.43 & 52.87 & 25.41 & 34.33 \\
\multicolumn{2}{l|}{}                                            & 500 & 66.43 & 65.03 & 65.72 & 74.88 & 73.08 & 73.97 & 84.07 & 78.82 & 81.36 & 49.00 & 27.07 & 34.88 \\ \hline
\multicolumn{2}{l|}{\multirow{10}{*}{\textbf{LLama3.1-8B}}}      & 0   & 43.33 & 36.56 & 39.66 & 46.18 & 30.53 & 36.76 & 38.44 & 51.39 & 43.98 & 11.83 & 17.13 & 14.00 \\
\multicolumn{2}{l|}{}                                            & 5   & 53.10 & 49.74 & 51.36 & 50.13 & 45.19 & 47.53 & 53.23 & 48.61 & 50.82 & 14.64 & 19.34 & 16.67 \\
\multicolumn{2}{l|}{}                                            & 10  & 55.05 & 48.86 & 51.77 & 56.72 & 50.72 & 53.55 & 60.65 & 58.33 & 59.47 & 15.85 & 24.86 & 19.35 \\
\multicolumn{2}{l|}{}                                            & 25  & 50.40 & 55.54 & 52.84 & 60.87 & 50.48 & 55.19 & 68.73 & 61.81 & 65.08 & 32.04 & 18.23 & 23.24 \\
\multicolumn{2}{l|}{}                                            & 50  & 60.00 & 47.45 & 52.99 & 61.68 & 54.57 & 57.91 & 64.79 & 76.04 & 69.97 & 21.68 & 27.07 & 24.08 \\
\multicolumn{2}{l|}{}                                            & 100 & 54.66 & 51.49 & 53.03 & 68.48 & 54.33 & 60.59 & 78.00 & 67.71 & 72.49 & 39.42 & 22.65 & 28.77 \\
\multicolumn{2}{l|}{}                                            & 200 & 60.82 & 54.83 & 57.67 & 67.40 & 59.13 & 63.00 & 75.60 & 76.39 & 75.99 & 35.00 & 27.07 & 30.53 \\
\multicolumn{2}{l|}{}                                            & 300 & 67.77 & 53.95 & 60.08 & 65.19 & 67.07 & 66.11 & 75.69 & 75.69 & 75.69 & 30.30 & 33.15 & 31.66 \\
\multicolumn{2}{l|}{}                                            & 400 & 66.67 & 59.26 & 62.75 & 67.88 & 71.63 & 69.71 & 78.25 & 77.43 & 77.84 & 31.34 & 34.81 & 32.98 \\
\multicolumn{2}{l|}{}                                            & 500 & 68.63 & 61.40 & 64.81 & 67.89 & 71.15 & 69.48 & 80.71 & 78.47 & 79.58 & 28.19 & 40.33 & 33.18 \\ \hline
\multicolumn{2}{l|}{\multirow{9}{*}{\textbf{LLama3.1-8B}}}       & 5   & 52.78 & 45.24 & 48.72 & 54.77 & 53.85 & 54.30 & 56.60 & 62.50 & 59.41 & 15.79 & 24.86 & 19.31 \\
\multicolumn{2}{l|}{}                                            & 10  & 58.01 & 58.52 & 58.27 & 63.66 & 52.64 & 57.63 & 59.17 & 69.44 & 63.90 & 21.62 & 22.10 & 21.86 \\
\multicolumn{2}{l|}{}                                            & 25  & 64.97 & 58.35 & 61.48 & 66.86 & 55.77 & 60.81 & 73.98 & 63.19 & 68.16 & 22.39 & 24.86 & 23.56 \\
\multicolumn{2}{l|}{}                                            & 50  & 62.57 & 61.69 & 62.12 & 67.78 & 58.65 & 62.89 & 77.95 & 68.75 & 73.06 & 27.33 & 24.31 & 25.73 \\
\multicolumn{2}{l|}{}                                            & 100 & 64.49 & 60.63 & 62.50 & 67.35 & 62.26 & 64.71 & 75.60 & 76.39 & 75.99 & 29.11 & 25.41 & 27.14 \\
\multicolumn{2}{l|}{}                                            & 200 & 65.70 & 59.93 & 62.68 & 69.81 & 64.91 & 67.27 & 77.05 & 78.13 & 77.59 & 38.14 & 24.86 & 30.10 \\
\multicolumn{2}{l|}{}                                            & 300 & 63.46 & 63.80 & 63.63 & 70.91 & 66.10 & 68.42 & 75.51 & 77.08 & 76.29 & 35.29 & 29.83 & 32.34 \\
\multicolumn{2}{l|}{}                                            & 400 & 65.17 & 63.80 & 64.48 & 67.88 & 71.63 & 69.71 & 77.21 & 78.82 & 78.01 & 33.15 & 33.70 & 33.42 \\
\multicolumn{2}{l|}{}                                            & 500 & 67.35 & 61.11 & 64.08 & 70.63 & 72.84 & 71.72 & 81.85 & 79.86 & 80.84 & 33.33 & 36.46 & 34.83 \\ \hline
\end{tabular}
}
\caption{Comprehensive many-shot ICL study of seven LLMs on four datasets.}
\label{tab:large-table}
\end{table*}

\section{Additional Details of ICA Framework} \label{apx:ica}

\subsection{More Detail of Annotation Refinement Methods} \label{apx:ica_refinement}

In this section, we provide detailed descriptions of the three refinement strategies used to enhance the quality of LLM-generated annotations: self-consistency \S \ref{apx:self_consistency}, self-correction \S \ref{apx:self_correction}, and error-aware refinement \S\ref{apx:error_aware}.

\subsubsection{Self-consistency} \label{apx:self_consistency}

To improve the robustness of LLM-generated annotations, we adopt a self-consistency refinement strategy inspired by prior work on decoding via multiple sampling \cite{wangself, xie-etal-2024-self}.  
In our NER setting, we apply self-consistency to low-confidence samples as identified by the confidence scoring method in Section~\ref{sec:ica_framework}.

For each selected sentence, we prompt the LLM to generate three independent predictions by randomly varying the order of in-context demonstration examples in the prompt.  
This results in three separate XML-style annotations for the same sentence.

We then perform majority voting over the predicted entities at both the span and type levels:  
\begin{itemize}
  \item \textbf{Span-level voting:} A candidate span is retained only if it appears in at least two out of the three predictions.
  \item \textbf{Type-level voting:} For spans that appear in multiple predictions, i.e., appear at least two times, the entity type assigned by the majority is selected as the final label.
\end{itemize}

This voting process filters out inconsistent or spurious predictions and prioritizes entities with high agreement across multiple completions.  
The final refined annotation includes only those entity mentions that meet the self-consistency threshold at both the span and type level.

\subsubsection{Self-correction} \label{apx:self_correction}

Self-correction is a lightweight refinement strategy that leverages the LLM’s own reasoning ability to revise and improve its initial predictions \cite{pan-etal-2024-automatically, kamoi-etal-2024-llms, heng-etal-2024-proggen, madaan2023self}. 
Given an initial annotation from ICA, we construct a second prompt that asks the LLM to review its output and make corrections if any errors are detected.

To help the LLM learn this behavior, we employ a few-shot prompting strategy.
Specifically, we begin by randomly sampling 5 sentences from the 100 human-labeled seed examples and holding them out.
We then use the LLM to generate predictions for the remaining 95 sentences.
From these predictions, we identify 6 examples that contain typical NER errors, two each for \textbf{spurious entity}, \textbf{missing entity}, and \textbf{type error}, as well as 4 examples that are correctly annotated.
Each selected example is paired with its corrected version and included in the prompt as a demonstration.
The other predictions are discarded and not used for prompting.

\begin{figure}[h]
    \centering
    \includegraphics[width=0.95\linewidth]{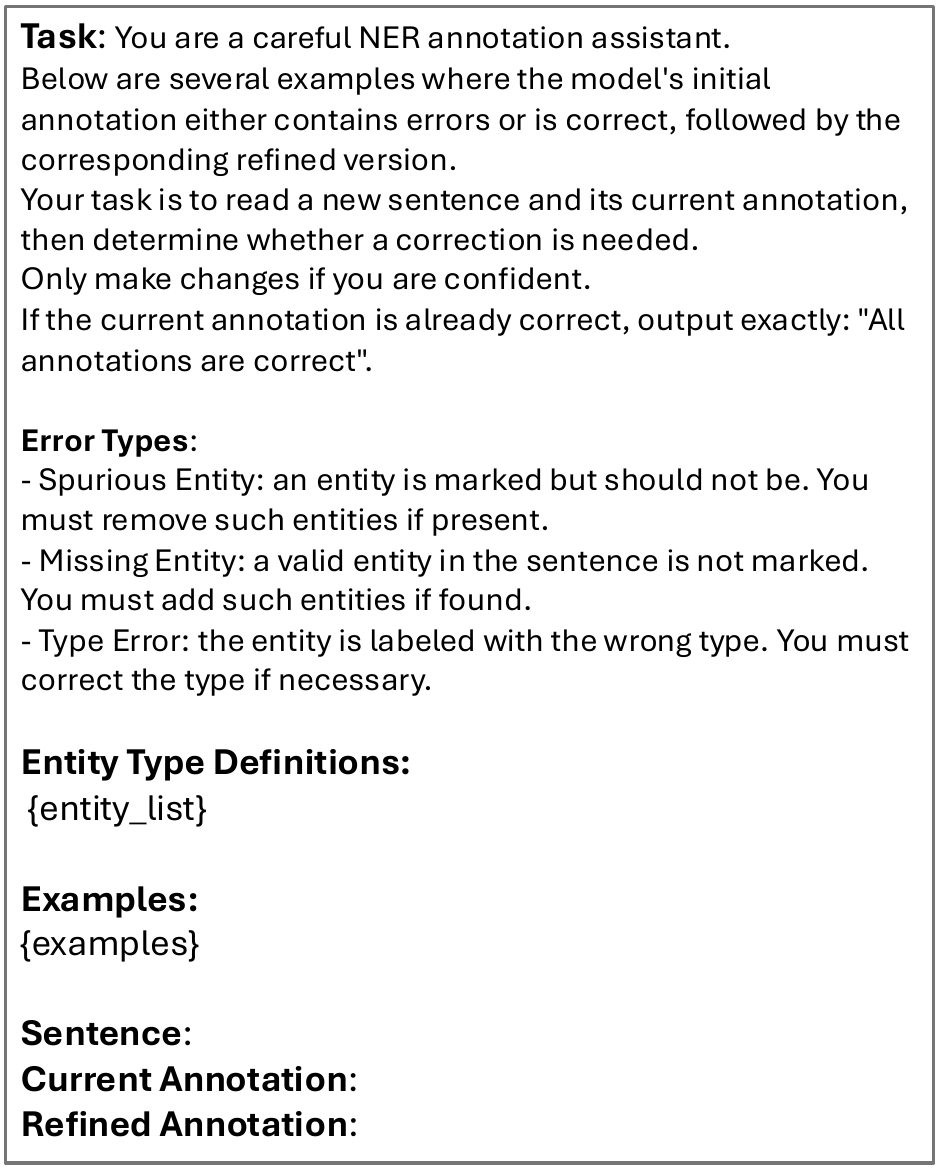}
\caption{Illustration of our self-correction prompt format.}
    \label{fig:self_correction_prompt}\end{figure}

Figure~\ref{fig:self_correction_prompt} shows the structure of our self-correction prompt, which consists of four parts:  
a brief description of common NER error types, entity type definitions, 10 few-shot examples (6 erroneous and 4 correct), and the target sentence with its current and refined annotations.

We do not explicitly target span boundary errors, as partially correct entity spans have been shown to be useful for training NER models \cite{mayhew-etal-2019-named}.  

\subsubsection{Error-Aware Refinement} \label{apx:error_aware}

Error-aware refinement (EAR) is a targeted strategy that applies a series of error-specific prompts to improve the quality of LLM-generated annotations.  
Unlike self-correction (Appendix~\ref{apx:self_correction}), which uses a single prompt to handle all types of errors, EAR decomposes the refinement process into three independent steps, each addressing one specific error type: spurious entities, missing entities, or incorrect types.

Each step is handled using a dedicated prompt, and the three prompts are applied sequentially to the same annotated sentence.  
For example, we first remove spurious entities, then add missing ones, and finally correct entity types.  
This modular setup improves controllability, avoids conflicting operations, and enables precise error handling.

To help the LLM learn how to correct specific error types, we follow the same few-shot example construction strategy used in the self-correction setup.  
We identify 10 examples from LLM-generated predictions—six containing typical NER errors (two per error type) and four that are correctly annotated.  
Each selected example is paired with its corrected version to form a demonstration pair.

To construct error-specific prompts, we further process these examples to isolate the target error type: for each prompt, we retain only the target error in the example and pre-correct all other types of errors within the same sentence.  
This ensures that each demonstration focuses solely on the intended error type, enabling the LLM to learn a precise and unambiguous correction behavior.

Unlike self-correction, where all examples are combined into a single prompt, EAR builds three separate prompts—each containing only examples of one specific error type.  
This design allows the LLM to focus on a single correction behavior at a time, without being distracted by unrelated errors.

\begin{figure}[h]
    \centering
    \includegraphics[width=0.95\linewidth]{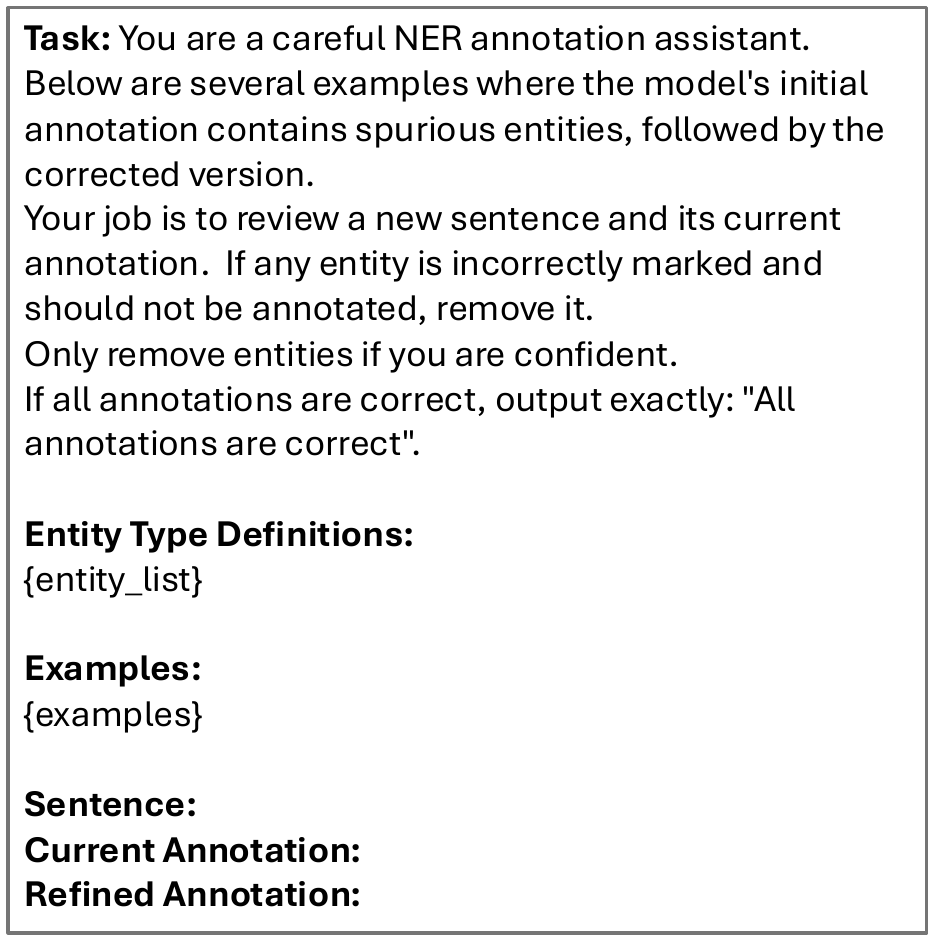}
    \caption{Prompt format for spurious entity error refinement.}
    \label{fig:ear_spurious}
\end{figure}

\begin{figure}[h]
    \centering
    \includegraphics[width=0.95\linewidth]{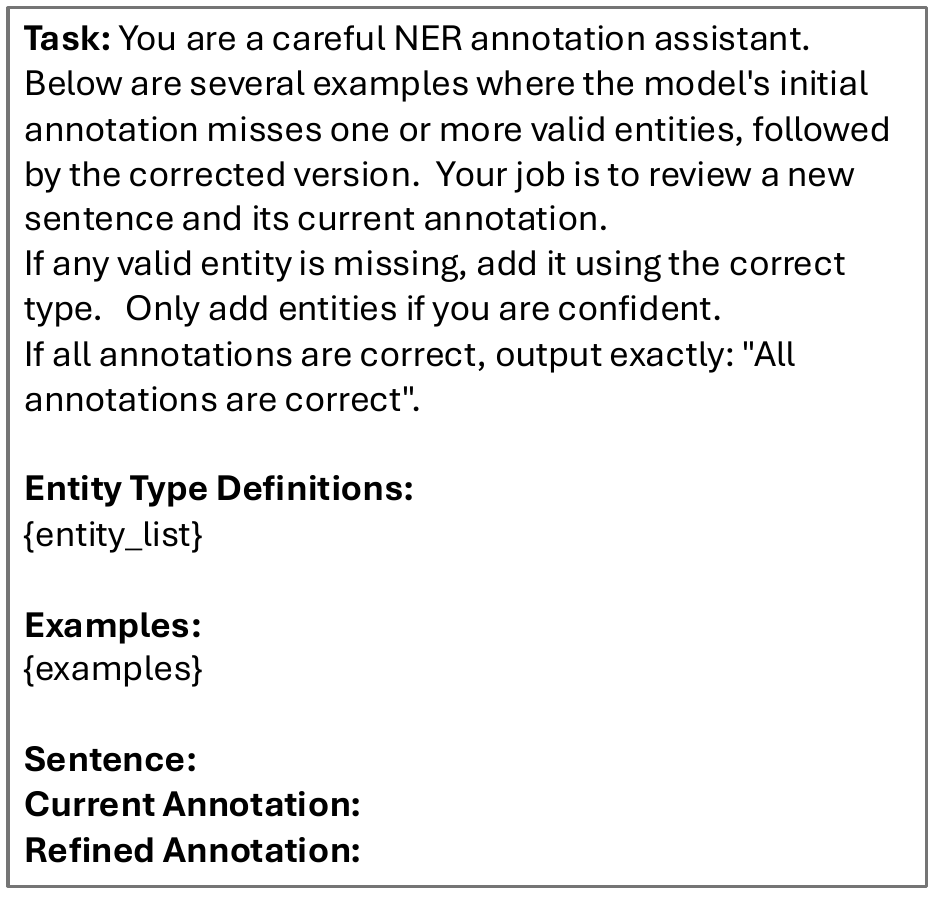}
    \caption{Prompt format for missing entity recovery.}
    \label{fig:ear_missing}
\end{figure}

\begin{figure}[h]
    \centering
    \includegraphics[width=0.95\linewidth]{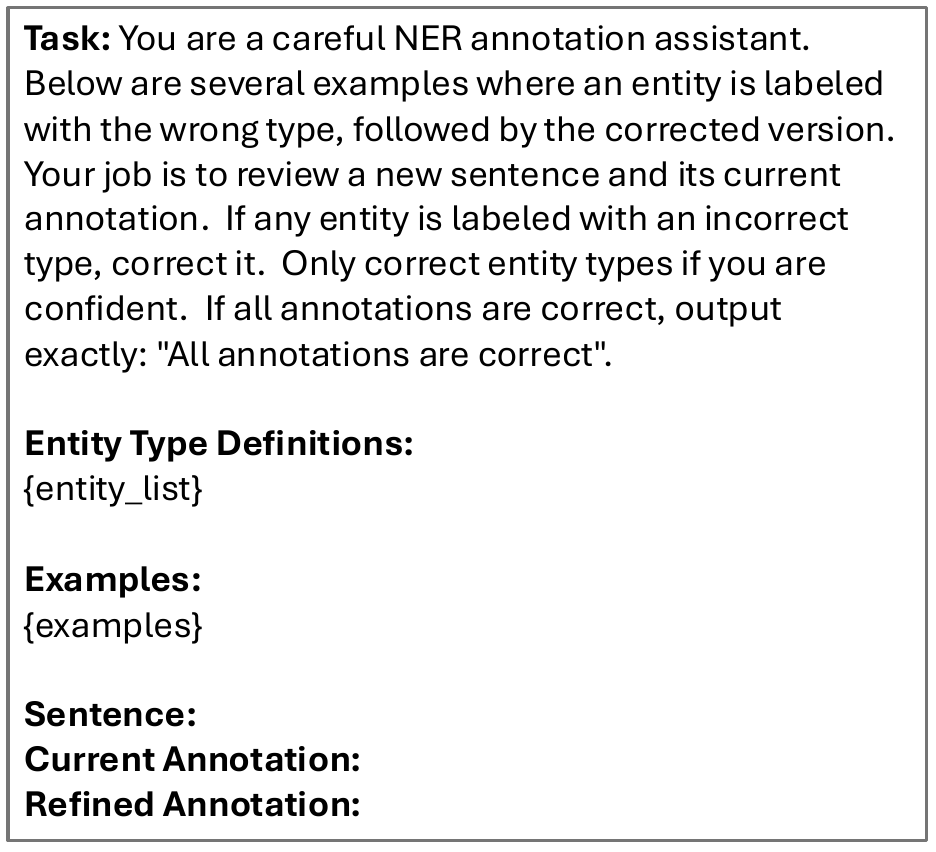}
    \caption{Prompt format for type error correction.}
    \label{fig:ear_type}
\end{figure}

Figures~\ref{fig:ear_spurious}, \ref{fig:ear_missing}, and \ref{fig:ear_type} illustrate the structure of the three error-specific prompts used in this step-by-step refinement process.

\subsection{CrossNER benchmark} \label{apx:corssner}

Table \ref{table:crossner_dataset} provided the detail statisitcs of each domain in CrossNER benchmark.

\begin{table*}[]
\renewcommand{\arraystretch}{1.15}
\centering
\resizebox{0.968\textwidth}{!}{
\begin{tabular}{c|c|c|c|c|c}
\hline
\multirow{2}{*}{\textbf{Domain}}                                  & \textbf{Unlabeled Corpus} & \multicolumn{3}{c|}{\textbf{Labeled NER}} & \multirow{2}{*}{\textbf{Entity Categories}}    \\ \cline{2-5}
   & \textbf{\# sentence}    & \textbf{\# Train}         & \textbf{\# Dev}         & \textbf{\# Test}        & \\ \hline
Reuters & -  & 14,987 & 3,466 & 3,684 & person, organization, location, miscellaneous \\ \hline
Politics     & 9.07M     & 200           & 541         & 651         & \begin{tabular}[c]{@{}c@{}}politician, person, organization, political party, event, \\ election, country, location, miscellaneous\end{tabular}    \\ \hline
\begin{tabular}[c]{@{}c@{}}Natural\\ Science\end{tabular}    & 5.32M     & 200           & 450         & 543         & \begin{tabular}[c]{@{}c@{}}scientist, person, university, organization, country, location, discipline, \\ enzyme, protein, chemical compound, chemical element, event,\\  astronomical object, academic journal,  award, theory, miscellaneous\end{tabular} \\ \hline
Music       & 9.82M      & 100           & 380         & 456         & \begin{tabular}[c]{@{}c@{}}music genre, song, band, album, musical artist, musical instrument,\\ award, event, country, location, organization, person, miscellaneous\end{tabular}     \\ \hline
Literature     & 9.17M         & 100           & 400         & 416         & \begin{tabular}[c]{@{}c@{}}book, writer, award, poem, event, magazine, person, location,\\  organization, country, miscellaneous\end{tabular}    \\ \hline
\begin{tabular}[c]{@{}c@{}}Artificial\\ Intelligence\end{tabular}   & 287.62K        & 100           & 350         & 431         & \begin{tabular}[c]{@{}c@{}}field, task, product, algorithm, researcher, metrics, university, \\ country, person, organization, location, miscellaneous\end{tabular}     \\ \hline
\end{tabular}
}
\caption{Data statistics of unlabeled domain corpora, labeled NER samples and entity categories for each domain.}
\label{table:crossner_dataset}
\end{table*}

\subsection{Case Study} \label{apx:case_study}

\begin{figure}[h]
    \centering
        \includegraphics[width=0.9\linewidth]{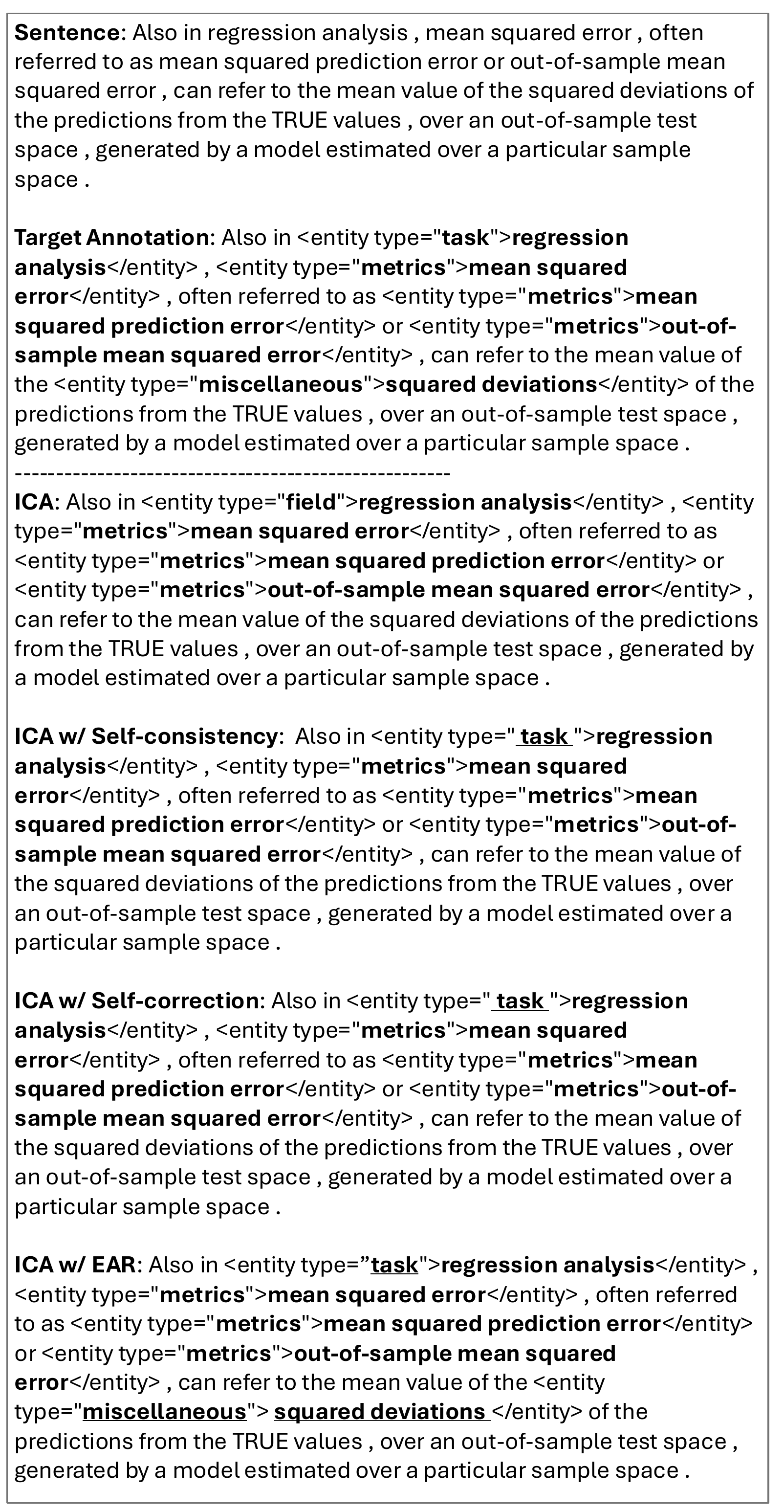}
    \caption{Case study from the AI dataset illustrating how the self-refinement methods correct annotation, where underline means the refinement method correct the error successfully.}
    \label{fig:case1}
\end{figure}

\begin{figure}[h]
    \centering
        \includegraphics[width=0.9\linewidth]{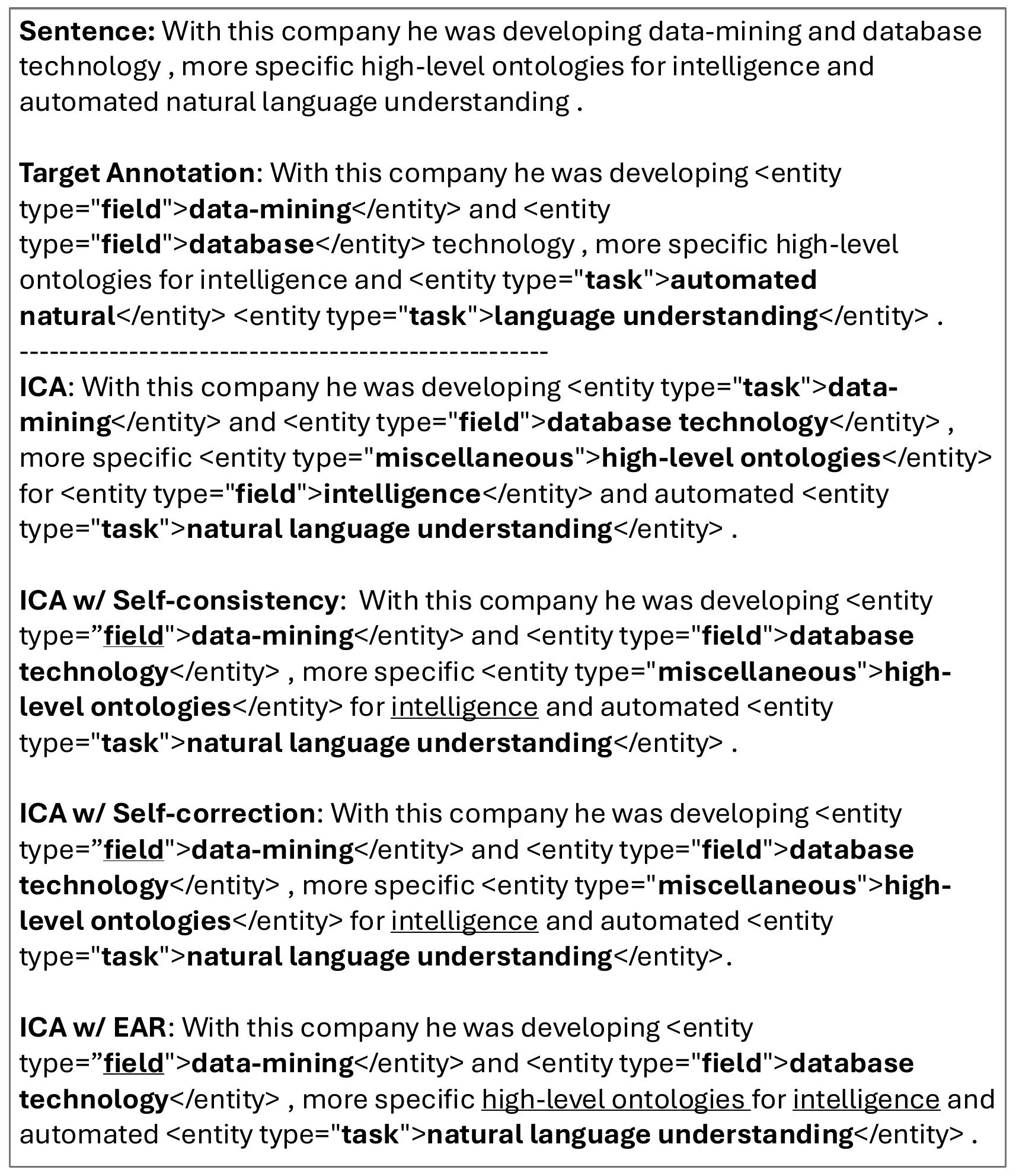}
    \caption{Case study from the AI dataset illustrating how the self-refinement methods correct annotation.}
    \label{fig:case2}
\end{figure}

To better understand the effect of our proposed self-refinement methods, we conduct a qualitative analysis by reviewing several annotated samples from the AI domains.
As illustrative examples, we present two representative cases in Figures~\ref{fig:case1} and~\ref{fig:case2}.

In Figure~\ref{fig:case1}, we observe that the initial ICA annotation incorrectly classifies 11regression analysis'' as a \texttt{field} entity, whereas it should be labeled as a \texttt{task}.
All three refinement strategies successfully correct this type error.
Additionally, EAR identifies an extra missing entity—``squared deviations''—and correctly labels it as \texttt{miscellaneous}, which was not detected by ICA.

In Figure~\ref{fig:case2}, all three methods correct the misclassification of ``data-mining'', which ICA originally labeled incorrectly, assigning it the correct type \texttt{field}.
They also successfully remove ``intelligence'', which was a spurious entity in the original annotation.
Moreover, EAR further removes ``high-level ontologies'', another spurious entity mistakenly included by ICA, demonstrating its stronger ability to filter over-predicted spans.

\subsection{Qualitative Observations and Design Motivations}
\label{sec:qualitative_challenges}
Beyond the quantitative results, we encountered several recurring challenges during the development of our ICA framework. These observations directly motivated the design choices behind our Error-Aware Refinement (EAR) strategy.

\paragraph{Sycophancy in Generic Self-Correction.} Our initial attempts used a generic prompt asking the LLM to ``check and fix any errors.'' We observed that the model frequently over-corrected its own outputs---deleting valid entities or altering correct types simply because it was prompted to look for mistakes---resulting in a notable drop in recall. This behavior is consistent with recent findings on the limitations of intrinsic self-correction~\citep{huang2023large, hao2025understanding}. 
The instability of open-ended correction directly motivated our decision to decompose refinement into three specific, directed sub-tasks in EAR (spurious, missing, type), where each prompt focuses on a single, well-defined error pattern.

\paragraph{Over-Sensitivity to Capitalization.} We found that LLMs tend to be over-sensitive to surface-level features such as capitalization. Non-entity capitalized phrases (e.g., ``Table 1,'' ``Section 3,'' or generic references like ``The University'') were frequently tagged as named entities. This failure mode was persistent across model sizes and motivated the dedicated Spurious-Entity step in our pipeline, which specifically targets this type of false positive.

\paragraph{Boundary Ambiguity in Specialized Domains.} In the Science and Literature domains, models struggled with precise span boundaries---for example, whether to include determiners or long modifiers in a book title or a chemical compound name. We found that expanding the number of in-context demonstrations mitigated this issue more effectively than adjusting zero-shot instructions, which aligns with our main finding on the benefit of scaling demonstrations.

\subsection{Cost Analysis and Deployment Scalability of ICA} \label{apx:cost_study}

Here we provide additional Analysis regarding the inference cost and scalability of our proposed ICA method. 
As explicitly discussed in Section~\ref{apx:ica}, our framework leverages LLMs exclusively as offline annotators rather than as real-time inference engines. 
The annotation process involving LLMs is a one-time cost incurred offline. 
After this initial annotation phase, subsequent inference and practical deployment rely solely on compact and efficient small language models (e.g., BERT), significantly reducing inference costs and enabling highly scalable deployments.

To further illustrate this benefit, we provide a concrete cost analysis based on reported annotation costs from prior literature. According to \cite{ding-etal-2023-gpt}, human annotation for the AI domain incurs a cost of \$42.85 per 100 samples. 
Consequently, the previous state-of-the-art method (DTrans-MPrompt), which requires 450 human-labeled samples for optimal performance (F1 score of 70.13), incurs a total annotation cost of \$192.74. In contrast, our ICA method achieves a higher F1 score (76.12) using only 100 human-labeled samples combined with 1,000 LLM-annotated samples. 
This setup results in a significantly reduced total annotation cost of only \$46.98 (human annotation: \$42.85 for 100 samples; LLM annotation: \$0.413 per 100 samples, totaling \$4.13 for 1,000 samples).

This cost analysis highlights that the ICA framework substantially lowers annotation expenses while delivering superior performance. More importantly, by separating the costly LLM annotation from inference, our method directly addresses real-world deployment considerations regarding cost-effectiveness and scalability.

\subsection{Class-wise performance} \label{apx:class_wise}

\begin{table*}[t]
\centering
\resizebox{\textwidth}{!}{
\begin{tabular}{lccccccccccccccc}
\toprule
\textbf{label} & algorithm & conference & country & field & location & metrics & misc & organisation & person & product & programlang & researcher & task & university & Avg. \\
\midrule
\textbf{ICL-ZS}   & 0.5675 & 0.7194 & 0.7863 & 0.6781 & 0.0286 & 0.0493 & 0.1596 & 0.6292 & 0.4000 & 0.6213 & 0.7480 & 0.8328 & 0.5660 & 0.5882 & 0.5779 \\
\textbf{ICL-MS}   & 0.7739 & 0.8346 & 0.8892 & \textbf{0.8268} & 0.5600 & 0.7622 & 0.3387 & 0.8140 & 0.8108 & 0.6616 & 0.7286 & 0.9097 & 0.6661 & 0.8971 & 0.7403 \\
\textbf{ICA}      & 0.7754 & 0.8468 & 0.8372 & 0.8033 & 0.7438 & 0.8183 & 0.3938 & \textbf{0.8531} & 0.8750 & 0.6935 & 0.8052 & 0.9325 & 0.7110 & 0.9032 & 0.7625 \\
\textbf{ICA w/ EAR} & \textbf{0.8150} & \textbf{0.8562} & \textbf{0.8905} & 0.8200 & \textbf{0.8099} & \textbf{0.8480} & \textbf{0.4444} & 0.8526 & \textbf{0.9038} & \textbf{0.7493} & \textbf{0.8235} & \textbf{0.9414} & \textbf{0.7467} & \textbf{0.9467} & \textbf{0.7964} \\
\bottomrule
\end{tabular}
}
\caption{Per-class performance on the AI domain.}
\label{table:per_class}
\end{table*}

Table~\ref{table:per_class} reports per-entity-type F1 across 14 AI-domain categories. ICA w/ EAR attains the best score on 12/14 types; the two exceptions are field (where ICL-MS is slightly higher) and organisation (where ICA is marginally ahead). On average, ICA w/ EAR reaches 0.7964 F1—an absolute gain of +0.056 over ICL-MS (0.7403; +7.6\% relative) and +0.034 over ICA (0.7625; +4.4\% relative). The largest margins (vs. ICL-MS) occur on long-tail or ambiguity-prone categories—location (+0.25), metrics (+0.09), misc (+0.11), task (+0.08), person (+0.09), and product (+0.09)—indicating improved robustness beyond head entities. We attribute these gains to EAR’s targeted repairs of missing, spurious, and typing errors (prompts and settings in the appendix). Overall, the improvements span both proper-noun types (e.g., person, organisation, country) and technical/semantic types (e.g., algorithm, metrics, task), supporting the generality of our approach rather than gains concentrated in a narrow subset of labels.

\section{AI Assistants In Writing} \label{apx:ai_assistant}

We utilized ChatGPT to identify writing issues and polish the manuscript for clarity and conciseness.

\end{document}